\pdfoutput=1

\documentclass[11pt]{article}

\usepackage{authblk}
\usepackage[]{acl}

\usepackage{times}
\usepackage{latexsym}
\usepackage{graphicx}
\usepackage{amssymb}

\usepackage{physics,amsmath}
\usepackage{cleveref}
\usepackage{enumitem}
\usepackage{booktabs}

\usepackage{textcomp, xspace}
\usepackage{pbox}
\usepackage{array}
\usepackage{multirow}
\usepackage{xcolor}
\usepackage[normalem]{ulem}
\useunder{\uline}{\ul}{}

\crefformat{section}{\S#2#1#3} 
\crefformat{subsection}{\S#2#1#3}
\crefformat{subsubsection}{\S#2#1#3}

\usepackage[T1]{fontenc}

\usepackage[utf8]{inputenc}

\usepackage{microtype}
\usepackage{arydshln}

\usepackage{inconsolata}

\newcommand{\model}{\textsc{MOKA}\xspace}
\newcommand{\data}{\textsc{Moral Events}\xspace}
\newcommand{\pretraindata}{\textsc{Morality Bank}\xspace}
\newcommand{\eqdef}{\mathrel{\mathop:}=}

\newcommand{\nop}[1]{}

\newcommand{\reda}[1]{\colorbox{magenta!10}{#1}}
\newcommand{\redb}[1]{\colorbox{magenta!25}{#1}}
\newcommand{\redc}[1]{\colorbox{magenta!45}{#1}}
\newcommand{\redd}[1]{\colorbox{magenta!70}{#1}}
\newcommand{\rede}[1]{\colorbox{magenta!100}{#1}}

\newcommand{\blue}[1]{\colorbox{cyan!15}{#1}}
\newcommand{\green}[1]{\colorbox{green!15}{#1}}
\newcommand{\yellow}[1]{\colorbox{yellow!80}{#1}}

\title{\model: Moral Knowledge Augmentation for Moral Event Extraction}

\author[1]{\textbf{Xinliang Frederick Zhang}}
\author[2]{\textbf{Winston Wu}}
\author[3]{\textbf{Nick Beauchamp}}
\author[1]{\textbf{Lu Wang}}

\affil[1]{Computer Science and Engineering, University of Michigan, Ann Arbor, MI}
\affil[2]{Department of Computer Science, University of Hawaii at Hilo, Hilo, HI}
\affil[3]{Department of Political Science, Northeastern University, Boston, MA}

\affil[ ]{$^1$\{\texttt{xlfzhang,wangluxy\}@umich.edu}}
\affil[ ]{$^2$\texttt{wswu@hawaii.edu}, $^3$\texttt{n.beauchamp@northeastern.edu}}

\begin{document}
\maketitle

\begin{abstract}

News media often strive to minimize explicit moral language in news articles, yet most articles are dense with moral values as expressed through the reported events themselves. However, values that are reflected
in the intricate dynamics among \textit{participating entities} and \textit{moral events} are far more challenging for most NLP systems to detect, including LLMs. To study this phenomenon, we annotate a new dataset, {\data}\footnote{Our data and codebase are available at \url{https://github.com/launchnlp/MOKA}.}, consisting of $5,494$ structured event annotations on $474$ news articles by diverse US media across the political spectrum. We further propose {\model}, a moral event extraction framework with MOral Knowledge Augmentation, which leverages knowledge derived from moral words and moral scenarios to produce structural representations of morality-bearing events. Experiments show that \model outperforms competitive baselines across three moral event understanding tasks.
Further analysis shows even ostensibly nonpartisan media engage in the selective reporting of moral events.

\end{abstract}
\section{Introduction}
Many news media frame their stories to further a particular ideological viewpoint \citep{scheufele1999framing}, often employing moral values rather than explicitly partisan language to subtly affect readers \citep{Haidt2007WhenMO,haidt2009above,lakoff2010moral,feinberg2015gulf}. However, existing NLP methods, including LLMs, face significant challenges in
discerning moral values. Past work has shown that these limitations may be due to lack of context \citep{Graham2009LiberalsAC,frimer2019moral}, lack of moral reasoning capabilities \citep{jiang2021can}, and the complexity of moral stances \citep{Zhou2023EthicalCC, Krgel2023ChatGPTsIM}. Detecting moral values is even harder for non-partisan news outlets which deliberately avoid explicit moral language in their reporting, but may express moral values indirectly 
by selecting which morally-laden events to report. Thus there remains an imperative need for NLP tools that can \textit{decipher} moral values latent in narrated events and interactions among entities.

 \begin{figure}[t!]
    \centering
    \includegraphics[width=0.48\textwidth]{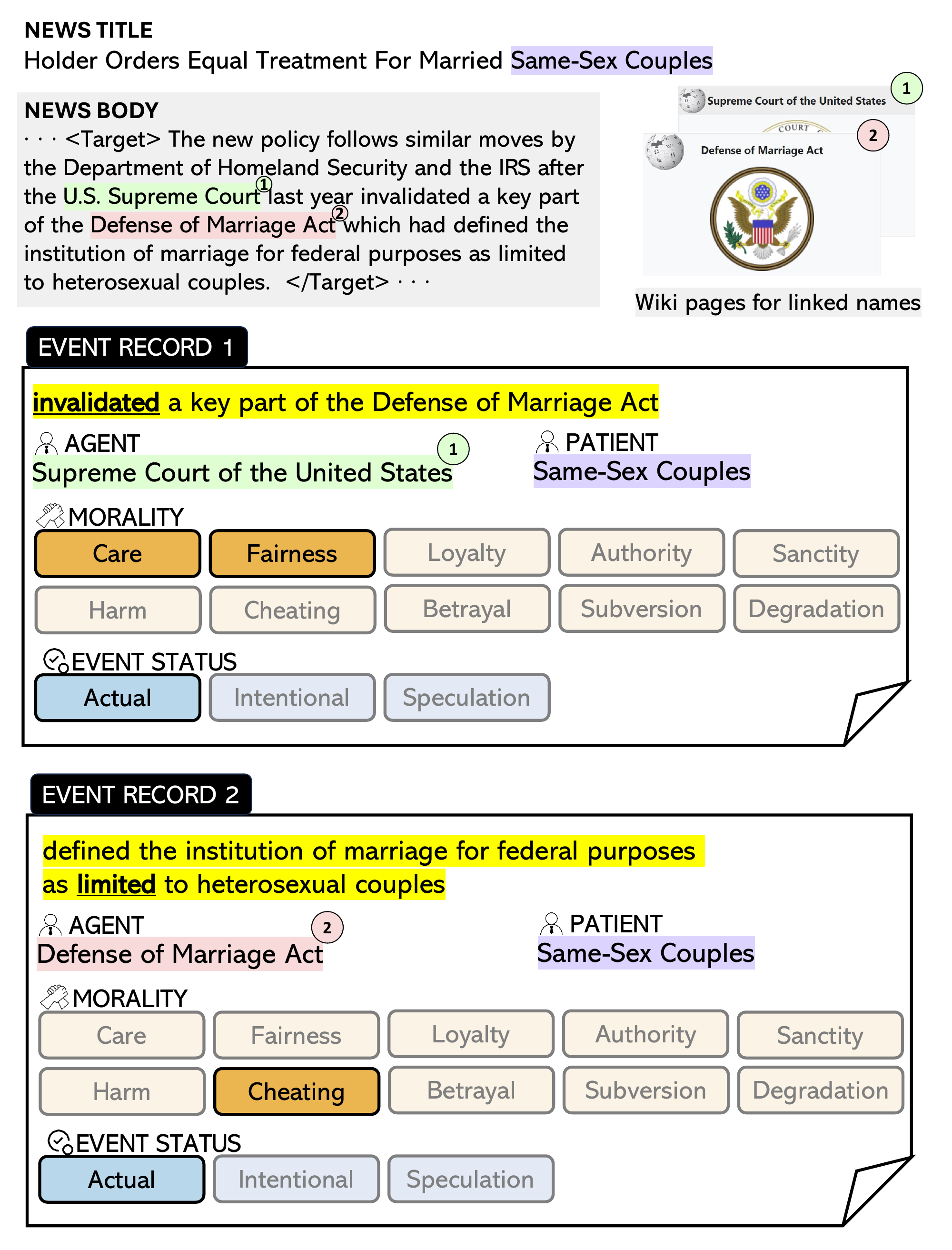}
    \vspace{-8mm}
    \caption{
    Sample moral event extractions (MEEs) for a target sentence from our \data dataset. Event participants are annotated per Wikipedia pages if applicable. In each event record, the {\ul\textbf{event trigger}} is a single word in an \yellow{event span}, and it might embody multiple moralities. 
    Moral event extraction is challenging due to several reasons: implicit participants (e.g. same-sex couples in Event Record 1) may not be mentioned in the target sentence, and understanding the relations among the participants is necessary to correctly infer the morality.
    }
     \vspace{-6mm}
    \label{intro_example}
\end{figure}

In particular, news articles tell complex stories that contain multiple people and events along with interactions among them. 
The participants in the events, the ordering of them, and the selection of events themselves have been shown to be useful for crafting impactful news articles~\citep{white2002media,van2013news,bourgeois2018selection}. 
In this work, we study morality and moral reasoning at the \textbf{event} level, enabling fine-grained structural analysis, capturing the nuances of relationships between participants performing moral actions, and uncovering the deeper layers of ethical dimensions intrinsic to news narratives. 
To this end, we first propose the concept of \textbf{moral events}, which capture the interaction among moral participants, such as moral agents and moral patients \citep{Gray2009-GRAMTD}, as demonstrated in \Cref{intro_example}. 
We then study the problem of \textit{structured} \textbf{moral event extraction}, which enables fine-grained analysis of how the choice of events in news articles and the context in which the events occur together carry moral implications, form effective news stories, and sway readers' perceptions.

Following prior work, we employ Moral Foundations Theory \citep[MFT;][]{Haidt2007WhenMO, Graham2009LiberalsAC, graham2013moral}, which posits five moral foundations, each containing two polarities of virtue and vice, e.g., \texttt{Care/Harm}.
MFT has been widely used in analyzing both mainstream news~\citep{hopp2021extended} and social media content~\citep{lin2018acquiring, hoover2020moral, trager2022moral}, but often in superficial ways based on explicit moral language, without utilizing the larger context and external knowledge
to better understand interactions between participants in moral actions. 
To illustrate the challenges in moral event understanding, \Cref{intro_example} shows a sentence containing multiple events with moral values but little in the way of explicit moral language.
The correct identification of Event Record 1 must take into account a longer context (e.g., the title) and background knowledge (that the Defense of Marriage Act governs same-sex couples) to identify the patient who is affected by the \textit{invalidated} event. Identifying the morality of \textit{invalidated} requires knowing that although it usually carries a negative connotation and might imply \texttt{Harm} on a surface level, here it is actually beneficial to the patient and in fact embodies \texttt{Fairness} and \texttt{Care} towards \textit{Same-Sex Couples}.

Our paper makes the following contributions.
First, we define a new schema of \textbf{moral events}, grounded in MFT and linguistics.
We then propose \textbf{moral event extraction (MEE)}: given unstructured text, detect morality-bearing event triggers, extract participants, and infer embodied moralities.

Second, to solve MEE, we curate a large dataset, \textbf{\data}, consisting of moral event annotations of news articles from diverse US media outlets. 
This dataset is unique in that annotations are conducted on multiple news articles about the same story, allowing us to analyze differences in how news outlets of different ideologies report moral events.
Moral participant annotations go beyond surface mentions and syntactic constraints, capturing \textit{implicit participants} in moral actions. 

We propose \textbf{\model}, a generative framework for MEE with \textbf{MO}ral \textbf{K}nwoledge \textbf{A}ugmentation. Capitalizing on the recent success of retrieval augmentation~\citep{Lewis2020RetrievalAugmentedGF, fevry-etal-2020-entities, izacard_few-shot_2022}, \model integrates moral knowledge derived from varying granularities, moral words and moral scenarios. 
Additionally, to support \model pre-training, we crawl a bank of 344k morality-bearing examples,  \pretraindata, leveraging validated morality lexicons \citep{Graham2009LiberalsAC, frimer2019moral}.
Extensive experiments highlight the usefulness and robustness of \model over strong baselines, including SOTA event extraction models and ChatGPT (\textit{gpt-3.5-turbo}). 
The results show that external moral knowledge is essential for LMs to excel at MEE and ethics-related moral reasoning in general. 
Further analysis of moral event reporting in news reveals substantive findings, including (1) left-right asymmetries where \texttt{Right}-to-\texttt{Left} moral events are more prevalent than the reverse regardless of underlying moral values or outlet ideology, and (2) a tendency of centrist media to focus primarily on moral events enacted by right-leaning entities.

\section{Related Work}

\subsection{NLP Benchmarks for Morality}

Recent NLP research has seen a surge in interest focusing on morality, including moral norms, ethical judgment, and  social bias,
Most work is based on MFT \citep[][]{Haidt2007WhenMO, Graham2009LiberalsAC}, a  social psychology theory that posits five moral foundations, each with two polarities: 
\texttt{Care/Harm}, \texttt{Loyalty/Betrayal}, \texttt{Fairness/Cheating}, \texttt{Authority/Subversion} and \texttt{Sanctity/Degradation}.

Many recently annotated morality \textbf{datasets} are limited to social media text, including Twitter \citep{, johnson-goldwasser-2018-classification, hoover2020moral, wang2021moral}
and Reddit \citep{lourie2021scruples, alhassan-etal-2022-bad, trager2022moral}.
Others combine social media text with crowdsourced data to study morality-related topics such as offensiveness \citep{sap-etal-2020-social}, rules of thumbs \citep{forbes-etal-2020-social}, knowledge of ethics \citep{hendrycks2020aligning}, branching narratives \citep{emelin-etal-2021-moral}, and everyday-situation judgments \citep{jiang2021can}. 
Only a few existing works study morality in news articles, mainly at the word-level \citep{mokhberian2020moral} or topic-level \citep{fulgoni-etal-2016-empirical,shahid-etal-2020-detecting}. 
By contrast, we collect a high-quality corpus of moral events from a wide range of news sources, to support the study of how the \textit{interplay} of events and moralities is used to craft effective news articles.

Most similar to our work are morality frames \citep{roy-etal-2021-identifying} and the eMFD corpus \citep{hopp2021extended}.
Although morality frames also capture participants in moral actions, they do not account for implicit patients affected by the moral action, who are not mentioned in the text span. Meanwhile, eMFD only annotates text spans and their embodied moralities.
In contrast, our work contains fine-grained structured event annotations including participants and linguistic features.

\paragraph{Moral foundation prediction}  
is a task treated as categorical classification, accomplished by fine-tuning pre-trained language models \citep{lin2018acquiring, alhassan-etal-2022-bad}.
Recent works approach it {with} template-based natural language generation \citep{forbes-etal-2020-social, jiang2021can}.
{While existing work focuses on predicting a moral label at the context-agnostic word-level \citep{Graham2009LiberalsAC, frimer2019moral} or document-level \citep{haidt2009above,mokhberian2020moral},} our models extract fine-grained structured moral events using both the context where events occur and the external moral knowledge. This allows us to capture nuances of moral actions involving different participants, and to better understand the role morality plays in shaping news narratives.

\subsection{Event Extraction}
Our work follows a long line of research in event extraction (EE), including two key stages: event detection (ED) and event argument extraction (EAE).
ED is defined as identifying an event trigger that best describes an event, i.e., change of state \citep{chen-etal-2018-collective, lou-etal-2021-mlbinet, deng-etal-2021-ontoed}, while EAE has the goal of extracting a phrase from text that mentions an event-specific attribute labeled with a specific argument role \citep{du-cardie-2020-document, li-etal-2021-document, parekh2022geneva}.

ED is commonly modeled as sequence labeling \citep{li-etal-2021-document}, question answering \citep{du-cardie-2020-event}, or template-based conditional generation \citep{hsu-etal-2022-degree}.
For the more challenging EAE task, three major approaches have been developed:
sequence labeling \citep{chen-etal-2015-event, nguyen-etal-2016-joint-event, du-cardie-2020-document} where global features have been incorporated to constrain the inference \citep{lin-etal-2020-joint};
question answering \citep{du-cardie-2020-event, tong-etal-2022-docee}, where models incorporate ontology knowledge about argument roles;
and generative models for structured extraction \citep{li-etal-2021-document, lu-etal-2021-text2event,du-ji-2022-retrieval}.  
More recently, LLMs have been used for EAE \citep{Zhang2024ULTRAUL}, but with subpar performance compared with specialized systems \citep{Li2023EvaluatingCI, Han2023IsIE}.

Our work proposes a new understanding task, \textit{moral event extraction}, a two-stage EE task with a special focus on morality-bearing events.
Unlike conventional EE, where each event type has its own event schema, we define a universal schema for moral events grounded in MFT and linguistics.
To the best of our knowledge, we are also the first to explicitly model \textit{multi-granularity moral knowledge} for EE tasks, as well as moral reasoning and understanding in general.

\section{\data Curation}
\label{data}

\label{schema}
We define a new structured schema for a \textbf{moral event} which represents a moral action, visualized in \Cref{intro_example}.
A moral event consists of moral agents, moral patients, a morality-bearing event span and event trigger, embodied morality, and event status. {We list major concepts below, and refer readers to \Cref{event_schema} for full descriptions.}
A moral action is performed or enabled by \textbf{moral agents} and affects \textbf{moral patients}.
An \textbf{event trigger} is usually a single word within an \textbf{event span} that can best characterize a moral action.
This span embodies one or more \textbf{moralities} in MFT \citep{Gray2009-GRAMTD}.
Note that moral patients may be \textit{implicit}: they do not have to be mentioned in a \textit{target sentence}.
To study the linguistic phenomenon, we further annotate \textbf{event status} which describes the factuality of an event \citep{Saur2009FactBankAC, lee-etal-2015-event}, i.e., whether an event is \textit{actual}, \textit{intentional} or \textit{speculative} \citep{app12105209}.

\paragraph{Annotation Process.}
\label{annotation_process}

We create our dataset, \data, using the following process.
We first sample 87 news stories from \textit{SEESAW} \citep{seesaw}, where each story contains 3 articles on the same event but reported by media of different ideologies.
To supplement this set with recent news, we further collect a new set of news article triplets from AllSides.com focusing on important issues in 2021 and 2022, e.g., abortion, gun control, and public health. We extract text from these articles using Newspaper\footnote{\url{https://github.com/codelucas/newspaper/}}, 
and clean all articles by removing boilerplate text and embedded tweets. 

Next,  moral events are annotated by native English speakers (at least two for each article).
Each annotator has access to all three articles in a story to maintain a non-biased view. we list the major steps of annotating a single article, with a detailed annotation protocol in \Cref{annotation_guideline}.
\begin{itemize}[leftmargin=2em,itemsep=0em,topsep=1pt,parsep=1pt,partopsep=1pt]
    \item[1.] The annotator first reads an article and then identifies agency-bearing entities that are participants in moral events. An entity may be of type Person, Organization, Geo-Political, or Other.\footnote{\textit{Other} includes religions (e.g., People of Faith) and topics (e.g., Homeland Security) among others.} Entities are coded by their canonical names, i.e., the names listed in Wikipedia, For example, mentions of ``President Trump'' or ``Trump'' are coded as ``Donald Trump''.
    
    \item[2.] For each sentence, the annotator identifies moral events and their attributes following the event schema defined in \cref{event_schema}. 
    
    \item[3.] Finally, the annotator determines the 5-way ideological leaning of the article. 
\end{itemize}

After an article is annotated (i.e., first pass), we proceed to the second pass to improve the annotation quality.
Specifically, we employ two distinct approaches: (a) an article is \textbf{revised} by a second person who corrects existing annotations and adds missing ones; (b) a second person annotates the full article from scratch following the procedures above, and a third person 
\textbf{merges} and resolves annotation conflicts.
83\% of articles are revised with approach (a) while 17\% adopt approach (b).

We ensured the quality of the annotations at multiple steps in the collection process (\Cref{data_quality_control}).  Also see  \Cref{tab:agreement} for annotation agreements.

\paragraph{Statistics.}
\data include $474$ news articles from $158$ stories, published by 63 different media outlets (26 left, 18 center, and 19 right).
On average, each article contains $11.6$ event annotations.
The articles cover $38$ salient topics reported from 2012 to 2022, 
and includes $1,952$ distinct entities
(see Table~\ref{tbl:top_30_list}).
We annotate diverse \textbf{entity types}:  People ($62.4$\%), Organization ($20.4$\%), Geo-Political ($9.6$\%), and Others ($7.6$\%).

\section{The \model Models}

 \begin{figure}[t]
    \centering
    \includegraphics[width=0.5\textwidth]{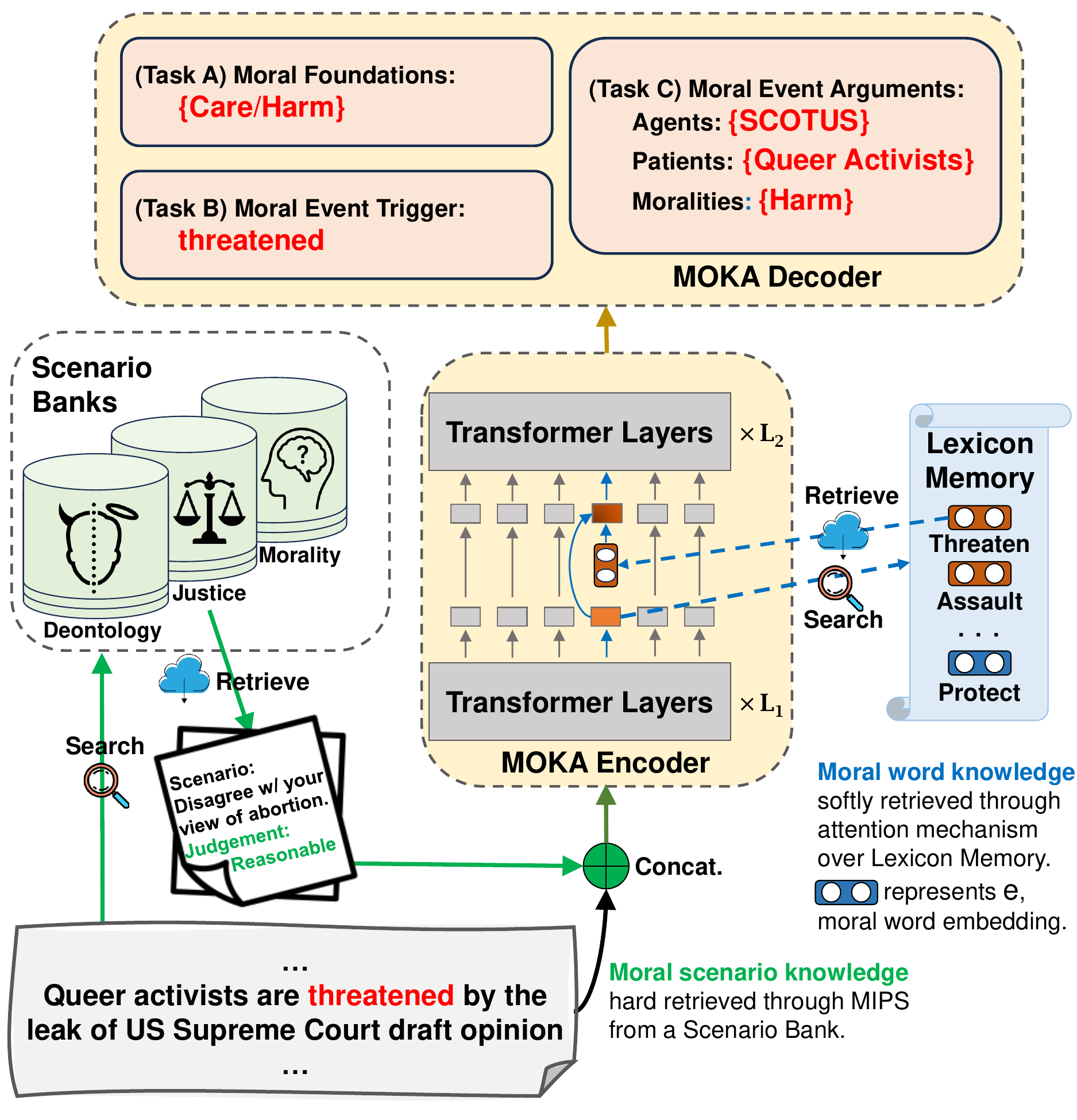}
    \vspace{-6mm}
    \caption{
    Overview of \model for (downstream) moral event extraction. 
    It highlights the process of retrieving and combining relevant \green{scenarios}, and the integration of moral \blue{word} knowledge through attention-based retrieval.  
    Embeddings in Lexicon are colored in \textcolor{red}{red} if moral words embody \texttt{Harm}, or \textcolor{blue}{blue} if \texttt{Care}.
     ``SCOTUS'' is an acronym for ``Supreme Court of the United States''. $\{\}$ indicates there can be multiple answers.
    }
    \vspace{-4mm}
    \label{fig:architecture}
\end{figure}

We now define the task of \textit{moral event extraction} (MEE), which extracts structured moral events from unstructured texts.
Similar to mainstream event extraction, we decompose MEE into the sub-tasks of event detection and event argument extraction, but with a focus on morality.
To tackle these tasks, we develop a new framework, \model (\Cref{fig:architecture}), which incorporates external moral knowledge into pre-trained language models at two levels: lexical-based moral word knowledge (\cref{moral_word}) and example-based moral scenario knowledge (\cref{moral_scenario}).
After two-stage pre-training, \model is fine-tuned on downstream tasks (\Cref{mee}).
We instantiate \model with \textit{Flan-T5-large}, though it is compatible with models of other architectures.

\paragraph{Moral Knowledge Augmentation.} 
To harness moral reasoning, it is critical to have a priori knowledge of necessary moral principles, just like a person of practical wisdom would \citep{ExplainingMoralKnowledge, practicalWisdom}. However, LLMs' access to moral facts is usually limited due to the lack of moral knowledge seen in the pretraining corpus \citep{jiang2021can}, although injecting morality into models has long been a question for debate \citep{wallach2008moral, awad2018moral}.

Models with a retrieval mechanism to access explicit non-parametric memory can provide provenance for their decision-making process and thus perform more robustly \citep{Lewis2020RetrievalAugmentedGF}.
So far, these retrieval mechanisms have been mostly used for certain knowledge-intensive tasks, such as entity-intensive question answering \citep{glass-etal-2022-re2g, chen-etal-2023-augmenting}. Hence, our work takes the first step to marry a retrieval component with moral knowledge to improve moral event understanding.

\subsection{Moral Word Knowledge}
\label{moral_word}

\paragraph{\textsc{Morality Bank} Construction.}
Unlike open domains where an existing knowledge base (KB) is always available such as the WikiData,
no such KB exists in the realm of moralities.
To start, we hypothesize that \textit{an utterance embodies a morality if it contains a morality-bearing mention}, where a moral mention is an occurrence of a moral word.\footnote{A moral word is a unique entry in the morality lexicon and the base form of moral mentions. Mentions like \textit{threatening} and \textit{threatened}, for example, both map to the \textit{Threaten} entry.}
We then combine two validated morality lexicons, MFD \citep{Graham2009LiberalsAC} and MFD2.0 \citep{frimer2019moral} into $891$ \textit{moral words}, and scrape example sentences that contain at least one moral mention from four authoritative online dictionaries.\footnote{Cambridge (UK \& US sites), Merriam-Webster, Dictionary.com, and YourDictionary.com.} We limit the sentence length to between 5 and 80 words, totaling $334k$ sentences. $95\%$ of example sentences are used for pre-training, and the rest for validation.
Samples are shown in Table~\ref{tbl:word_example}.

\paragraph{Lexicon Memory Access.} 
Similar to \citet{fevry-etal-2020-entities} and \citet{verga-etal-2021-adaptable}, we define the \textit{Lexicon Memory} $\mathbf{E}$ as a matrix containing one embedding for each moral word. For each word, we initialize and freeze its embedding, $\mathbf{e}$, by averaging the contextualized representations of its mentions in \textsc{Morality Bank}.

When encoding a sentence, a moral mention is first tagged with a special token pair \texttt{(<Morality>, </Morality>)}. 
For each mention, \model computes a query vector $\mathbf{h}_q$, which is the averaged representation of the special token pair and the enclosed moral mention. $\mathbf{h}_q$ is then used to retrieve relevant moral knowledge $\mathbf{h}_m$ from the Lexicon Memory via a single-head attention mechanism, $\mathbf{h}_m = \textrm{Attn}(\mathbf{h}_q,\mathbf{E})$, where Attn($\cdot$,$\cdot$) is the cross-attention mechanism in \citet{Vaswani2017AttentionIA}. 
Finally, the sum of $\mathbf{h}_m$ and $\mathbf{h}_q$ is normalized, and fed to the next layer.
Following \citet{fevry-etal-2020-entities}, we interleave standard transformer layers with the Lexicon Memory access layer at a lower layer, which is the $8^{\text{th}}$ layer ($L_1=8$ and $L_2=16$ in \Cref{fig:architecture}).

\vspace{-2mm}
\paragraph{Moral Word Knowledge Pre-training.} 
The pretraining objective is a combination of language modeling ($\mathcal{L}_{LM}$), morality prediction ($\mathcal{L}_{MV}$), moral word linking ($\mathcal{L}_{MWL}$), and moral label association ($\mathcal{L}_{MLA}$), each described below.
\noindent \textbf{Language modeling} is employed to train \model to denoise corrupted sentences, to familiarize itself with moral language usage. 
\textbf{Morality prediction} is introduced to provide a direct signal to train \model to uncover the morality(s) embodied in a morality-bearing input sentence. To prevent \model from learning shortcuts, the \textit{seed word} used to scrape the input sentence is always masked.
Two new training objectives are also proposed to train the memory access mechanism effectively. For each moral mention, the \textbf{moral word linking} objective guides \model to identify the corresponding moral word by learning to maximize the attention score over the correct entry, e.g., \textit{Threaten} in Figure~\ref{fig:architecture}.
The \textbf{moral label association} objective promotes \model's capability of associating a moral mention and its embodied morality(s). It is achieved by, for each morality embodied by a mention, maximizing the summation of attention scores over all moral words in $\mathbf{E}$ that share the same morality. To handle moral words that are associated with multiple moralities, we use \textit{multi-label margin loss} (\cref{appx:eq:lmla}).
Compared to cross entropy, this objective flattens scores over target moralities and mitigates saturated gradients. Detailed mathematical formulations of Lexicon Memory Access are in \Cref{appx:lma}.

Our work differs from existing work using entity memory 
\citep{fevry-etal-2020-entities, verga-etal-2021-adaptable} in three aspects. 
First, moral concepts and stances are more abstract than concrete entities. No KB exists in the context of morality, so we curate \pretraindata, transforming morality-bearing sentences into a structured knowledge base. In addition, unlike entity memory which can utilize entity-linking tools out-of-the-box, we rely on designed objectives $\mathcal{L}_{MWL}$ and $\mathcal{L}_{MLA}$ to enable memory access.

\subsection{Moral Scenario Knowledge}
\label{moral_scenario}

\noindent \textbf{Moral Scenario Bank Compilation.}
While fundamental theoretical moral theories are prescriptive and rule-based, we depart from this approach and adopt example-based, descriptive moral scenarios. As pointed out by \citet{jiang2021can}, while human can directly understand abstract moral principles without the need for interaction with concrete moral scenarios, those principles are too perplexing for machines. Thus, we guide \model to develop its moral sense by immersing itself in real-world moral scenarios. To achieve this, we compile a suite of Moral Scenario Banks by incorporating three large-scale ethics-related datasets: \texttt{Delphi} \citep{jiang2021can}, \texttt{Social Chemistry} \citep{forbes-etal-2020-social}, and \texttt{ETHICS} \citep{hendrycks2020aligning}. We convert them into 7 moral scenario banks, with statistics and examples in \Cref{tbl:scenario_example}.

\noindent \textbf{Scenario Retrieval.}
For each moral scenario bank, we convert (scenario, label) pairs into key-value pairs. Then, we encode all keys into dense vectors using the Flan-T5 encoder, which ensures an isomorphic embedding space between searching and reasoning.
To implement efficient maximum inner-product search (MIPS), we create a ScaNN index \citep{avq_2020} and search top-$K$ ($K=3$ in this study) relevant scenario pairs using dot product similarity between the query and keys, i.e., scenarios.
The retrieved scenario pairs are concatenated together with the input, which is then fed into \model encoder, as shown in Figure~\ref{fig:architecture}.

\noindent \textbf{Moral Scenario Knowledge Pre-training.}
We pre-train \model on moral scenario banks to improve its moral reasoning by guiding it to navigate the complex interplay of diverse moral principles within real-world scenarios. The task is formulated as: given an input \textit{scenario} and a set of relevant scenario pairs in (scenario, label) format,
e.g., (``enjoying your life with your family'', ``morally good''), \model should generate a desired output.

To help further digest the retrieved scenarios and enhance the encoder's moral reasoning capabilities, 
we introduce a new pre-training objective -- Retrieved Label Masking (RLM). Specifically, we randomly mask out the label of one retrieved example and apply MLM objective to recover this label. By explicitly training the encoder to discern the associated moral label, it helps \model from collapsing to simply memorizing retrieved labels and making trivial inferences.

This approach is in line with retrieval augmented generation, where existing work mainly empowers a language model with a retriever to fetch text-form knowledge items from an external knowledge bank \citep{Lewis2020RetrievalAugmentedGF, fan-etal-2021-augmenting, shi-etal-2022-nearest, izacard_few-shot_2022}. 
Most existing work is limited to the use of a single knowledge source, with the exception of \citet{kic2023pan, zhang-etal-2023-merging}. On the contrary, \model embraces multiple moral knowledge sources under different moral scenarios. 

\subsection{Downstream Moral Event Extraction}
\label{mee}
Figure~\ref{fig:architecture} depicts the flow of \model with \textbf{dual knowledge augmentation} for downstream moral event extraction (MEE) tasks. Concretely, the input passage is first used to retrieve $K$-scenarios pairs ($K$=3) from the moral scenario bank on which \model is pre-trained. Retrieved scenarios are combined with the original input to form a moral knowledge-enriched input. Next, we tag moral mentions on the fly, and follow the Lexcion Memory Access steps outlined in \Cref{moral_word} to integrate moral word knowledge. \model is then trained to generate an end-task-specific output with three training objectives:   $\mathcal{L}_{\text{FT}} = \mathcal{L}_{CE} + \mathcal{L}_{MWL} + \mathcal{L}_{MLA}$, where $\mathcal{L}_{CE}$ is a standard cross-entropy loss applied to the decoder, and $\mathcal{L}_{MWL}$ and $\mathcal{L}_{MLA}$ are the same memory-access losses as described in \Cref{moral_word}. Meanwhile, as presented in Table~\ref{tbl:result_task_A} and \ref{tbl:result_task_BC}, we disable $\mathcal{L}_{MLA}$ if we want to examine MOKA’s efficacy when not explicitly informed of the specific morality(s) embodied by each moral mention.

For \model variants with single knowledge augmentation, we remove the corresponding module on which the variant is not pre-trained. For example, for \model augmented with moral scenario knowledge only, moral mention tagging and moral word knowledge integration (\cref{moral_word}) are not applied.

\begin{table}[t]
\centering
\resizebox{0.95\linewidth}{!}{%
\begin{tabular}{lcccc}
\toprule
\multicolumn{1}{c}{\multirow{2}{*}{Model}}                                                     & \multicolumn{2}{c}{\data}                    & \multicolumn{2}{c}{eMFD Corpus} \\
\cmidrule(lr){2-3} 
\cmidrule(lr){4-5}
\multicolumn{1}{c}{}                                                                           & F1                   & Acc.                 & F1             & Acc.           \\ \midrule
\multicolumn{5}{l}{\textbf{Baselines}}                                                                                                                                         \\
Dictionary-based counting (\citeauthor{brady2017emotion})                                                                        & 45.8                 & 56.8                 & 33.0           & 52.0           \\
RoBERTa (large; \citeauthor{Liu2019RoBERTaAR})                                                                                  & 63.6                 & 82.6                 & 28.7           & 69.0           \\
POLITICS (base; \citeauthor{politics})                                                                                      & 62.7                 & 82.4                 & 29.0           & 68.8           \\
ChatGPT (zero-shot; \citeauthor{Li2023EvaluatingCI})                                                                            & 41.2                 & 69.9                 & 31.9           & 66.9           \\
ChatGPT (few-shot; \citeauthor{Li2023EvaluatingCI})                                                                             & 46.9                 & 75.6                 & 30.5           & 69.1           \\
Flan-T5 (large; \citeauthor{Chung2022ScalingIL})                                                                                & 62.0                 & 83.6                 & 25.4           & 68.4           \\ \midrule
\multicolumn{5}{l}{\textbf{MOKA with moral word knowledge augmentation only}}                                                                                                  \\
Pretrain on \texttt{Morality Bank} only                                                                            & 63.6                 & \reda{83.9}                 & 27.3           & 69.0           \\
\;\; + moral word linking  ($\mathcal{L}_{MWL}$)                                      & \reda{63.9}                 & \reda{83.9}                 & 27.8           & 69.0           \\
\;\;\;\; + moral label association ($\mathcal{L}_{MLA}$) & \redb{64.0}                 & \reda{83.9}                 & 28.5           & 69.1           \\ \midrule
\multicolumn{5}{l}{\textbf{MOKA with moral scenario knowledge augmentation only}}                                                                                              \\
Delphi (moral judgement; \citeauthor{jiang2021can})                                                                       & \reda{63.7}                 & \redb{84.1}                 & 30.4           & \redb{70.4}           \\
\;\; + RLM                                                      & 62.3                 & \reda{83.8}                 & 30.1           & \redb{70.3}           \\
Deontology  (\citeauthor{hendrycks2020aligning})                                                                                   & 62.5                 & 83.6                 & 30.5           & \redc{70.5}           \\
\;\; + RLM                                                      & 62.2                 & 83.5                 & 30.4           & \redb{70.4}           \\
Social chem (foundation; \citeauthor{forbes-etal-2020-social})                                                                        & 62.2                 & \reda{83.7}                 & 32.4           & \redc{70.6}           \\
\;\; + RLM                                                      & \redb{64.1}                 & \reda{84.0}                 & 32.5           & \redc{70.7}           \\ \midrule
\multicolumn{5}{l}{\textbf{MOKA with dual moral knowledge augmentation}}                                                                                                  \\
Delphi (moral judgement; \citeauthor{jiang2021can})                                                                       & 63.3                 & 83.6                 & 32.9           & \redc{70.7}           \\
\;\; - $\mathcal{L}_{MLA}$                                 & \reda{63.9}                 & \redb{84.1}                 & 32.1           & \redc{70.6}           \\
Deontology   (\citeauthor{hendrycks2020aligning})                                                                                    & \redb{64.0}                 & \reda{84.0}                 & 32.9           & \redc{70.8}           \\
\;\; - $\mathcal{L}_{MLA}$                                  & \redb{64.2}                 & \reda{84.0}                 & \redc{\textbf{34.3}}  & \redc{\textbf{71.1}}  \\
Social chem (foundation; \citeauthor{forbes-etal-2020-social})                                                                        & \redc{\textbf{65.3}}        & \redb{\textbf{84.3}}        & \redc{33.7}           & \redc{71.0}           \\
\;\; - $\mathcal{L}_{MLA}$                                  & \redb{64.1} & \reda{84.0} & \redb{33.4}           & \redc{71.0}           \\ \midrule
Improvements over best baseline                                                                & 2.7\%                & 0.8\%                & 3.9\%          & 2.9\%   \\
\bottomrule
\end{tabular}
}
\vspace{-2mm}
\caption{
Weighted F1 and accuracy on \data and eMFD \citep{hopp2021extended} for Task A  (average of 5 runs). 
Best results are in \textbf{bold}. \model{s} that outperform all baselines are highlighted on a scale of 5 red shades. 
``+'' and ``-'' indicate the inclusion or exclusion of a particular training objective.
{\ul \model augmented with dual moral knowledge (RLM enabled) achieve better performances across the board by notable margins.} 
Full results and color scheme explanations are in Table~\ref{tbl:full_result_task_A}. 
}
\vspace{-4mm}
\label{tbl:result_task_A}
\end{table}

\section{Experiments}

\subsection{Tasks and Datasets}
\label{sec:tasks}

\begin{table*}[t]
\centering
\resizebox{0.75\linewidth}{!}{%
\begin{tabular}{lcccccc}
\toprule
\multicolumn{1}{c}{\multirow{2}{*}{Model}}                                                     & \multicolumn{1}{c}{Task B} & \multicolumn{5}{c}{Task C}                                                    \\ 
\cmidrule(lr){2-2} 
\cmidrule(lr){3-7}
\multicolumn{1}{c}{}                                                                           & Trigger EM                 & Morality F1      & Agent EM      & Agent F1      & Patient EM    & Patient F1    \\ \midrule
\multicolumn{7}{l}{\textbf{Baselines}}                                                                                                                                                                               \\
DEGREE (base; \citeauthor{hsu-etal-2022-degree})                                                                                  & 45.5                       & 53.0          & 47.3          & 58.6          & 30.1          & 39.2          \\
DEGREE (large; \citeauthor{hsu-etal-2022-degree})                                                                                 & 46.2                       & 54.2          & 49.2          & 60.3          & 30.5          & 40.3          \\
ChatGPT (zero-shot; \citeauthor{Li2023EvaluatingCI})                                                                            & 19.5                       & 39.5          & 30.3          & 49.8          & 12.3          & 23.2          \\
ChatGPT (few-shot; \citeauthor{Li2023EvaluatingCI})                                                                             & 32.1                       & 38.1          & 34.2          & 51.4          & 20.1          & 30.6          \\
Flan-T5 (large; \citeauthor{Chung2022ScalingIL})                                                                                & 46.2                       & 53.8          & 47.5          & 59.4          & 30.8          & 41.2          \\ \midrule
\multicolumn{7}{l}{\textbf{MOKA with moral word knowledge augmentation only}}                                                                                                                               \\
Pretrain on \texttt{Morality} \texttt{Bank} only                                                                            & 45.3                       & \redb{54.6}          & 47.5          & 59.9          & \redb{31.2}          & \redb{41.7}          \\
\;\; + moral word linking  ($\mathcal{L}_{MWL}$)                   & 45.6                       & \redc{55.9}          & 47.6          & 59.8          & \redc{\textbf{31.5}} & \redb{41.7}          \\
\;\;\;\; + moral label association ($\mathcal{L}_{MLA}$) & 46.2                       & \redd{57.0}          & 48.3          & 60.2          & \redb{31.3}          & \redb{41.9}          \\ \midrule
\multicolumn{7}{l}{\textbf{MOKA with moral scenario knowledge augmentation only}}                                                                                                                           \\
Delphi (moral judgement; \citeauthor{jiang2021can})                                                                       & \redb{47.0}                       & \redd{57.5}          & 48.5          & \reda{60.4}          & \reda{30.9}          & \reda{41.4}          \\
\;\; + RLM                                                                                          & \redc{47.4}                       & \redc{55.6}          & 48.5          & 60.3          & \redb{31.2}          & \redb{41.5}          \\
Deontology   (\citeauthor{hendrycks2020aligning})                                                                                   & 46.1                       & \redb{54.8}          & 49.0          & \redb{60.9}          & \reda{30.9}          & \redb{41.6}          \\
\;\; + RLM                                                                                          & \redc{47.2}                       & \redc{56.0}          & \redb{\textbf{49.5}} & \redb{61.2}          & \redb{31.3}          & \redc{\textbf{42.1}} \\
Social chem (foundation; \citeauthor{forbes-etal-2020-social})                                                                      & \redb{46.7}                       & \redd{56.5}         & 48.9          & \redb{\textbf{61.4}} & \redb{31.0}          & \reda{41.4}          \\
\;\; + RLM                                                                                          & \redc{47.5}                       & \redc{56.0}          & 48.8          & \reda{60.5}          & \redb{31.0}          & \redb{41.7}          \\ \midrule
\multicolumn{7}{l}{\textbf{MOKA with dual moral knowledge augmentation}}                                                                                                                               \\
Delphi (moral judgement; \citeauthor{jiang2021can})                                                                       & \redc{47.4}                       & \redd{56.8}          & 48.1          & 60.3          & 30.2          & 40.5          \\
\;\; - $\mathcal{L}_{MLA}$                                                                      & \redb{46.7}                       & \redd{57.2}          & 47.6          & 60.0          & 30.2          & 40.5          \\
Deontology  (\citeauthor{hendrycks2020aligning})                                                                                     & \redb{46.8}                       & \rede{\textbf{58.2}}         & 47.9          & 60.3          & \reda{30.9}          & 41.1          \\
\;\; - $\mathcal{L}_{MLA}$                                                                      & \redd{\textbf{48.1}}              & \redd{57.3}          & 48.2          & \redb{61.0}          & 30.7          & 41.1          \\
Social chem (foundation; \citeauthor{forbes-etal-2020-social})                                                                      & \redb{46.5}                       & \rede{58.1} & 48.4          & \redb{61.0}         & 30.5          & 40.8          \\
\;\; - $\mathcal{L}_{MLA}$                                                                      & \redb{46.7}                       & \redd{57.7}          & 48.2          & \reda{60.5}          & 30.0          & 40.1          \\ \midrule
Improvements over best baseline                                                                                  & 4.1\%                      & 7.4\%         & 0.6\%         & 1.8\%         & 2.3\%         & 2.2\%             \\
\bottomrule
\end{tabular}
}
\vspace{-2mm}
\caption{
Results on \data for Tasks B and C (average of 5 runs). 
Best results are in \textbf{bold}. \model{s} that outperform all baselines are highlighted on a scale of 5 red shades. 
``+'' and ``-'' indicate the inclusion or exclusion of a particular training objective.
{\ul \model augmented with dual knowledge achieve consistently better performances on trigger detection and morality inference, while the best results on participant extractions (i.e., \textit{agent} and \textit{patient}) are reached by single-knowledge variants.} 
Full results and color scheme explanations are in Table~\ref{tbl:full_result_task_BC}.
}
\vspace{-4mm}
\label{tbl:result_task_BC}
\end{table*}

We conduct holistic evaluations on three moral event extraction sub-tasks using two datasets: the newly curated \data and eMFD \citep{hopp2021extended}. The input in all tasks is a 4-sentence document which includes a \textit{target sentence}, a preceding and a succeeding sentence, and a title.

\noindent \textbf{Task A: Moral foundation prediction.} Conditioned on a document and \textit{one} moral event span, make a 5-way judgment on the moral foundation for the given moral event. 

\noindent \textbf{Task B: Moral event trigger detection.} Given a document, detect moral event triggers from the target sentence. 

\noindent \textbf{Task C: Moral event argument extraction.} Given a document and \textit{one} moral event span, produce triplets in the form of moral agents, patients, and a 10-way morality inference. This demands profound moral reasoning skills to correctly understand the interplay between participants and moralities.

As eMFD \citep{hopp2021extended} only annotates moral foundations but not  event attributes, it is only applicable to Task A. Also, since each document might embody more than one foundation or morality, we follow existing research on approaching multi-label classification with generative models \citep{yang-etal-2018-sgm, Yue2021CliniQG4QAGD, Chai2022PromptBasedGM} by consistently linearizing foundations or moralities as a sequence in our experiments.

We split \data by chronological order, and use the 90 news articles published in the 2nd half of 2022 as the test set. We sample a subset of articles from eMFD, and partition them randomly on the article level. 
\Cref{tbl:split_stats} shows the detailed statistics of splits on both datasets.

\subsection{Baselines and \model Variants}

For Task A, we follow \citet{alhassan-etal-2022-bad} and compare with encoder-only models: RoBERTa \citep{Liu2019RoBERTaAR} and its variant continually trained on news, POLITICS \citep{politics}. We include a dictionary approach \citep{brady2017emotion}, where the moral foundation is determined by the presence of moral words defined in morality lexicons \citep{Graham2009LiberalsAC, Buttrick2020HistoricalCI}.
\footnote{To prevent trivially predicting all foundations, we consider the top-3 moral foundations based on counting frequency.} 
For Task B and C, since they are newly introduced in this work to study different aspects of moral events, we follow the EE literature and compare with a SOTA baseline, DEGREE \citep{hsu-etal-2022-degree}.
For all tasks, we also compare with Flan-T5 \citep{Chung2022ScalingIL} with downstream fine-tuning only, and ChatGPT (\textit{gpt-3.5-turbo}).\footnote{Prompts, adapted from \citet{Li2023EvaluatingCI}, are shown in Table~\ref{tbl:chatgpt_prompt}.}

We consider three \textbf{\model variants}. First, with \textbf{moral word knowledge augmentation only}, we experiment with pretraining on \texttt{Morality Bank} only with  $\mathcal{L}_{LM}$ and $\mathcal{L}_{MV}$ objectives. We then incrementally add the new $\mathcal{L}_{MWL}$ and $\mathcal{L}_{MLA}$ objectives. For \textbf{moral scenario knowledge augmentation only}, we connect \model with one Moral Scenario Bank at a time,\footnote{We also experimented with conflating all Moral Scenario Banks together.  However, this did not improve performance.} and test the effectiveness of the RLM objective. Putting all together, we obtain the full model with the \textbf{dual moral knowledge augmentation}. We further examine \model's efficacy with and without $\mathcal{L}_{MLA}$ in the \textit{moral word knowledge pre-training stage}.

\subsection{Results}

\noindent \textbf{Evaluation Metrics.} 
We report accuracy and weighted F1 for moral foundation prediction in Task A and morality inference in Task C.
For trigger detection (Task B), We consider Trigger F1-score, the same criterion as in prior work \citep{wadden-etal-2019-entity, lin-etal-2020-joint}. 
For participants extraction (i.e., agents and patients) in Task C, we follow QA \citep{rajpurkar-etal-2016-squad, rajpurkar-etal-2018-know} and EE \citep{du-cardie-2020-document, tong-etal-2022-docee} communities, and adopt span-level Exact Match (EM) and token-level F1 as two evaluation metrics.  

Table~\ref{tbl:result_task_A} shows the results for Task A. Performances on \data and eMFD exhibit distinct trends. \data follows a natural moral foundation distribution (\Cref{tbl:distribution}), whereas eMFD has a roughly even distribution.
Encoder-only models show strong performances on both datasets, where RoBERTa-large achieves the best F1 scores on \data. ChatGPT, despite its stunning capability, struggles to understand and discern moral foundations. Flan-T5-large, the backbone model in \model, yields unsatisfying results, especially on eMFD, due to a lack of ethics-related documents in its pertaining stage \citep{jiang2021can}.
In contrast, moral knowledge augmentation in \model improves Flan-T5's moral reasoning capabilities by $35\%$ (F1 of 34.3 vs. 25.4). 

Table~\ref{tbl:result_task_BC} presents model results on Task B and C. Similar to Task A, ChatGPT performs worse than specialized EE systems.
While DEGREE is a SOTA model in the general domain,
it does not outperform fine-tuning a Flan-T5 model, highlighting the unique challenges posed by moral event understanding. 
On the other hand, when equipped with dual moral knowledge, \model yields the best results for trigger detection and morality inference. Particularly, the $7\%$ performance gain on morality inference can be attributed to \model effectively assimilating moral knowledge at different granularities after two stages of moral knowledge-centric pre-training.
Note, however, that single-knowledge variants reach the best participant extraction results, and our hypothesis is that the injected moral knowledge does not make participants available in moral reasoning.
{Further, as shown in \Cref{tab:event_status}, we study moral reasoning abilities across different event statuses: 1) Triggers of \textit{Actual} events are easier to detect than others; 2) Morality of \textit{speculative} events can be better identified, due to a higher usage of explicit moral language.}

\section{Further Analyses}

We further investigate the use of moral language in news media through the lens of selective reporting of moral events. We validate past work showing how different ideologies focus on different moralities (\textbf{RQ1}), and go beyond that to show how the selective reporting of moral \textit{narratives} 
reveals more subtle and asymmetrical forms of bias (\textbf{RQ2}).

\begin{figure}[t]
    \centering
    \includegraphics[width=0.42\textwidth]{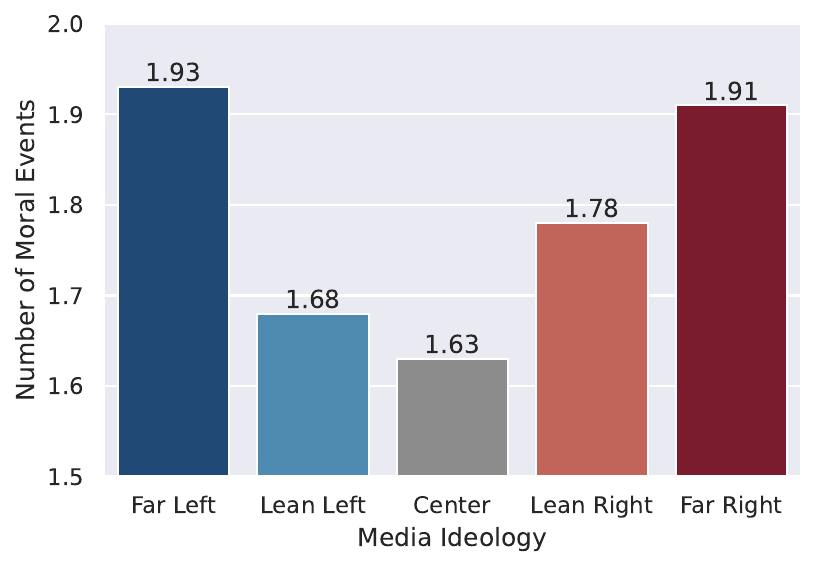}
    \vspace{-2mm}
    \caption{
    Number of moral events in each 100-word segment. Highly partisan media outlets tend to include more moral language than non-partisan ones.
    }
    \vspace{-6mm}
    \label{fig:analysis_0}
\end{figure}

\begin{figure}[t]
    \centering
    \includegraphics[width=0.45\textwidth]{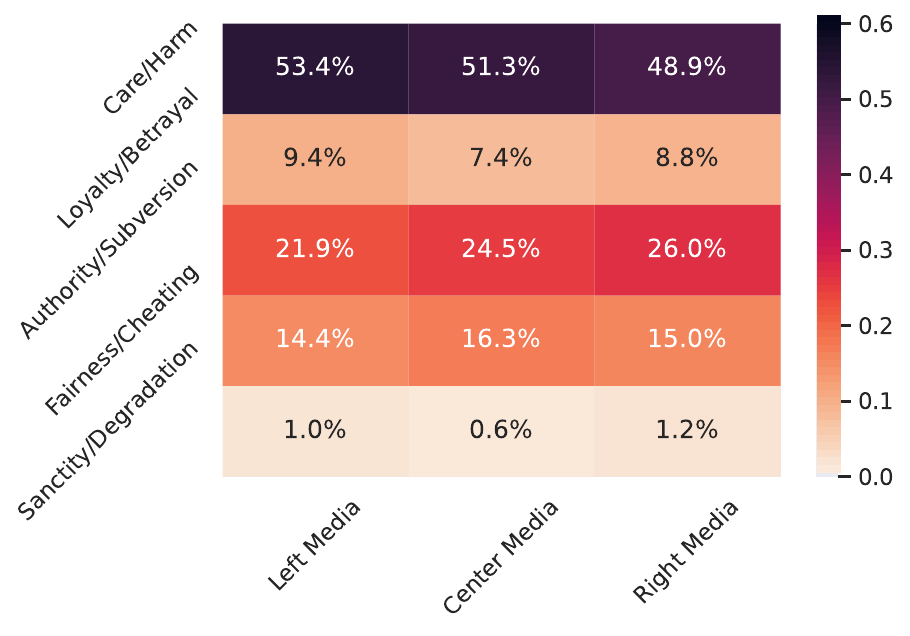}
    \vspace{-2mm}
    \caption{Employed moral foundation distribution by media outlets of different ideologies. 
    }
    \vspace{-6mm}
    \label{fig:analysis_1}
\end{figure}

\paragraph{RQ1: Does moral language usage correlate with media ideology?}
 Figure~\ref{fig:analysis_0} shows that more extreme outlets unsurprisingly tend to use more moral language overall, whereas the centrist media use it least. The most frequent moral foundation is \texttt{Care/Harm}, in line with findings on social media text \citep[Figure~\ref{fig:analysis_1};][]{hoover2020moral, trager2022moral}. Both news media and social media use relatively little \texttt{Sanctity/Degradation}, which measures religious purity and disgust and is rarely reported in the news. However, news media use a far higher proportion of \texttt{Authority/Subversion} than social media because much of the news focuses on politicians and other ruling figures. In contrast, social media covers more \texttt{Fairness/Cheating}, due to its greater focus on explicit morality as seen in AITA Reddit forum which has a special focus on personal ethical violations \citep{alhassan-etal-2022-bad}. Finally, within these general tendencies, we  also find that left-leaning media focus more on \texttt{Care/Harm}, while right-leaning media focus more on \texttt{Authority/Subversion}, in line with MFT \citep{Graham2009LiberalsAC}. 

\paragraph{RQ2: How is media bias revealed by the selective reporting of \textbf{agent-morality-patient} narratives?}

Moral narratives are fundamentally constructed out of three elements: an agent, a patient, and an action with an associated morality. To understand how ideology and morality shape the news, we must examine these three elements jointly.

To measure agent and patient ideologies, all entities that appear in at least two news articles were coded by a domain expert for their partisan leaning on a binary left/right scale, yielding $197$ coded entities and $1,253$ associated events. Figures~\ref{fig:analysis_2_care_harm} and~\ref{fig:analysis_2_aut_sub} show the correlations between agent-patient relationship and outlet ideologies for the two most prevalent foundations, \texttt{Care/Harm} and \texttt{Authority/Subversion}. This reveals rich differences between left, right, and center media that do not fall into the simple partisan symmetries that have been posited previously \citep{Gentzkow2005MediaBA, fbcd98f1-f03b-30c4-a611-956047a66b70, Graham2009LiberalsAC}.

Within \texttt{Care/Harm} (Figure~\ref{fig:analysis_2_care_harm}), the left media report relatively more \texttt{Right}-harm-\texttt{Left} events than the right media do, and vice versa. Interestingly, an asymmetry is observed that media across different ideologies all report more \texttt{Right}-harm-\texttt{Left} events than the reverse (i.e. \texttt{Left}-harm-\texttt{Right}). That applies to centrist outlets as well, which 
show a pronounced tendency of reporting more \texttt{Care/Harm} where the \texttt{Right} entity is the agent.

For \texttt{Authority/Subversion} (Figure~\ref{fig:analysis_2_aut_sub}), we find both left and right outlets report more \texttt{Authority} from \texttt{Right-to-Left}, while centrist media are once again more focused on \texttt{Right}-agent events overall. These asymmetries are even more notable with \texttt{Subversion}, where we see right media reporting (disapprovingly) on \texttt{Left} entities subverting the \texttt{Right} but also (approvingly) on \texttt{Right} subverting the \texttt{Left},
while centrist media also report more \texttt{Right}-subverts-\texttt{Left} events.

To summarize, mainstream media strive for balance in ideological language, entities, and even expressed values, but when we examine agent-value-patient triplets, ideological differences become evident. 
We found both important \textbf{left-right asymmetries}, and the \textbf{distinctive behavior of centrist media}, which overwhelmingly focuses on \texttt{Right} agents. These illustrate the importance of event-level morality analysis in political news.

\section{Conclusion}

We studied the task of moral event extraction---a novel reasoning task with the objective of, given unstructured text, producing structural representations for morality-bearing events including their triggers, participating entities, and embodied morality. 
To support this study, we curate a new dataset, \data, including $5,494$ structured annotations. We propose \model, a moral reasoning-enhanced event extraction framework with moral knowledge augmentation. Specifically, we employ retrieval augmentation by integrating moral knowledge at varying granularities, derived from moral words and moral scenarios. Further analyses reveal the effectiveness of using moral events to discern ideological biases even when outlets report seemingly objective events.

\section*{Acknowledgments}
This work is supported in part through National Science Foundation under grant IIS-2127747, Air Force Office of Scientific Research under grant FA9550-22-1-0099, and computational resources and services provided by Advanced Research Computing (ARC), a division of Information and Technology Services (ITS) at the University of Michigan, Ann Arbor. We appreciate ACL ARR reviewers for their helpful feedback. We thank Jahnavi Enaganti, Yunwoo Chang, Miya Brado, Syafawani Abdul Rahim and Natalie Wilson for their efforts in \data construction. 

\section*{Limitations}
\paragraph{GPU resources.}
The framework proposed in this work is an encoder-decoder based generative model. It is thus more time-consuming than standard discriminative models for training and evaluation, which in turn results
in a higher carbon footprint. Specifically, we train each model ($\sim770$ million parameters) on 1 single NVIDIA RTX A40 with significant CPU and memory resources. The training time for each model ranges from 1 to 3 days, depending on the configurations. 

\paragraph{\data and Annotations.}
While we offer comprehensive training guidelines and implement necessary quality control processes, users of our \data might not fully agree with our annotated structural annotations of moral events. We deeply respect different views, especially those from underrepresented groups, and are eager to explore variations in how individuals from different geographical backgrounds interpret these events in future work.

Due to budget constraints, we were only able to annotate $474$ articles and $5,494$ moral events. However, it is worth noting that, the annotated events in \data already outnumber one of the most prevalent event extraction dataset -- ACE 2005 \citep{ace2005}. Future endeavors might leverage  AI systems (e.g., ChatGPT) to scale up moral event annotations with minimal human efforts.  

\paragraph{Moral Foundation Theory.}
In this study, we build our approach \model on top of a prominent social psychology theory -- Moral Foundation Theory (MFT). MFT, however, has its own \textit{cultural bias}. That is, the theory is largely based on research conducted in Western cultures, particularly in the United States. As such, the concluded five moral dimensions might not be universally applicable. Furthermore, we assume a \textit{static nature} in MFT, i.e. there is a stable set of moral foundations. However, recent work embarked on splitting the dimension of \texttt{Fairness} into \texttt{Equality} and \texttt{Proportionality} \citep{atari2023morality}, and extending the original MFT to include \texttt{Liberty} \citep{iyer2012understanding} and \texttt{Honor} \citep{atari2020foundations}, which need to be taken into account in future modeling as well.

\section*{Ethical Consideration}

\noindent \textbf{\data collection.} All news articles were collected in a manner consistent with the terms of use of the original sources as well as the intellectual property and the privacy rights of the original authors of the texts, i.e., source owners. During data collection, the authors honored privacy rights of content creators, thus did not collect any sensitive information that can reveal their identities. All participants involved in the process have completed human subjects research training at their affiliated institutions. We also consulted Section 107\footnote{\url{https://www.copyright.gov/title17/92chap1.html\#107}.} of the U.S. Copyright Act and ensured that our collection action fell under the fair use category.

\smallskip
\noindent\textbf{\data annotation.} In this study, manual work is involved. All the participants are college students, and they are compensated fairly (15 USD/hr per school policy). We hold weekly meetings to give them timely feedbacks and grade them quite leniently to express our appreciation for their consistent efforts. Lastly, they consent that their annotated data can be further repurposed and distributed for research purposes.

\clearpage

\setcounter{table}{0}
\setcounter{figure}{0}
\renewcommand{\thefigure}{A\arabic{figure}}
\renewcommand{\thetable}{A\arabic{table}}


\clearpage
\appendix
\begin{figure}[t]
    \centering
    \includegraphics[width=0.45\textwidth]{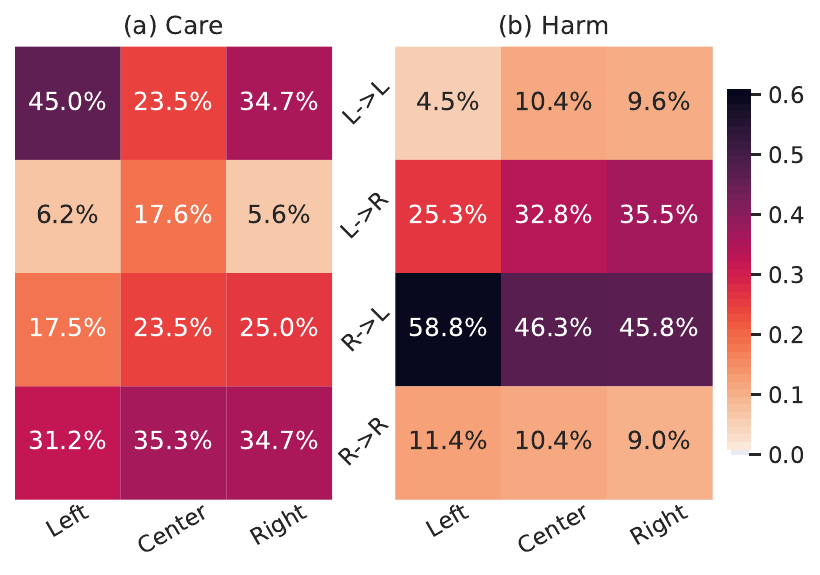}
    \vspace{-2mm}
    \caption{Correlation among agent-patient relationships, media outlet ideologies, and \texttt{Care}-/\texttt{Harm}-bearing moral events. Each percentage indicates the proportion of reporting a certain agent-patient interaction, and each column sums up to $100\%$. For example, $6.2\%$ means that, among all \texttt{Care}-bearing events reported by left-leaning media, $6.2\%$ of them are enabled by a \texttt{Left}-leaning entity and affecting a \texttt{Right}-leaning entity. 
    }
    \label{fig:analysis_2_care_harm}
    \vspace{-2mm}
\end{figure}

\begin{figure}[t]
    \centering
    \includegraphics[width=0.45\textwidth]{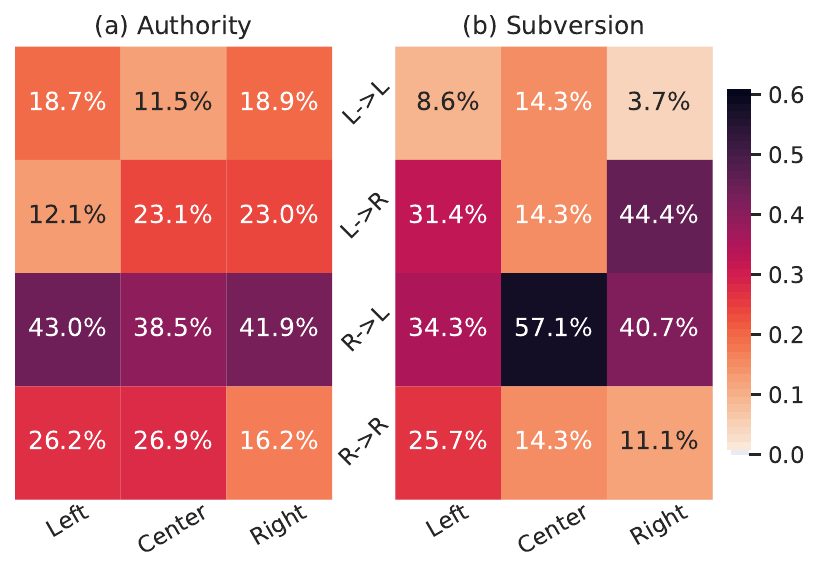}
    \vspace{-2mm}
    \caption{Correlation among agent-patient relationships, media outlet ideologies, and \texttt{Authority}-/\texttt{Subversion}-bearing moral events. Each percentage indicates the proportion of reporting a certain agent-patient interaction, and each column sums up to $100\%$.
    }
    \label{fig:analysis_2_aut_sub}
    \vspace{-2mm}
\end{figure}

\begin{table}[t]
    \centering
    \small
    \begin{tabular}{lll}
        \toprule
        Attribute & Merged & Revised \\
        \midrule
        Agent        & 0.77 & 0.94 \\
        Patient      & 0.64 & 0.92 \\
        Morality     & 0.67 & 0.92 \\
        Event Status & 0.59 & 0.91 \\
        \bottomrule
    \end{tabular}
    \caption{Krippendorff's alpha on various event attributes for revised (approach a) and merged (approach b) event annotations.}
    \vspace{-2mm}
    \label{tab:agreement}
\end{table}

\begin{table}[t]
\centering
\resizebox{0.85\linewidth}{!}{%
\begin{tabular}{lr}
\toprule
Entity                                       & Frequency \\ 
\midrule
Americans                                    & 156       \\
Donald Trump                                 & 123       \\
United States                                & 118       \\
Republican Party                             & 100        \\
Joe Biden                                    & 93        \\
Democratic Party                             & 87        \\
Barack Obama                                 & 81        \\
United States Congress                       & 72        \\
People                                       & 58        \\
Supreme Court of the United States           & 52        \\
Federal Government of the United States      & 45        \\
Justice Department                           & 41        \\
Biden Administration                         & 35        \\
Hillary Clinton                              & 27        \\
United States House of Representatives       & 27        \\
United States Senate                         & 24        \\
White House                                  & 23        \\
Immigrants                                   & 22        \\
Trump Administration                         & 22        \\
Obama Administration                         & 21        \\
Police                                       & 20        \\
Affordable Care Act                          & 20        \\
Women                                        & 20        \\
Federal Bureau of Investigation              & 18        \\
Ukraine                                      & 18        \\
Food and Drug Administration                 & 18        \\
Senate Republicans                           & 18        \\
State Department                             & 17        \\
Mitch McConnell                              & 17        \\
Lawmakers                                    & 16        \\\bottomrule
\end{tabular}
}
\caption{Top-30 frequent entities in \data sorted by their frequencies, i.e., the number of articles in which an entity appears.}
\label{tbl:top_30_list}
\end{table}

\begin{table*}[t]
\centering
\resizebox{0.8\linewidth}{!}{%
\begin{tabular}{lp{0.95\linewidth}} \toprule
\textbf{Sentence}    & Napoleon now realized that it would be impossible , without running serious risks , to {\ul \textless{}Morality\textgreater{}{} \textcolor{red}{oppose} \textless{}/Morality\textgreater{}{}}$_{\text{Subversion}}$ the movement in favor of {\ul \textless{}Morality\textgreater{}{} unity \textless{}/Morality\textgreater{}{}}$_{\text{Loyalty}}$ .                                                                                                                                                                                                                                                                                                                \\
\textbf{Morality}           &  Subversion       \\ \midrule
\textbf{Sentence}            & While waiting for emergency {\ul \textless{}Morality\textgreater{} help \textless{}/Morality\textgreater{}}$_{\text{Care}}$ to arrive , the {\ul\textless{}Morality\textgreater{} \textcolor{red}{victim} \textless{}/Morality\textgreater{}}$_{\text{Harm}}$ should wash the {\ul \textless{}Morality\textgreater{} wound \textless{}/Morality\textgreater{}}$_{\text{Harm}}$ site with soap and water and then keep the {\ul \textless{}Morality\textgreater{} injured \textless{}/Morality\textgreater{}}$_{\text{Harm}}$ area still and at a level lower than the heart .                 \\
\textbf{Morality}            & Harm                                                                                                                                                                                                                                                                                                                                                                                                                                                                                                                                                                                           \\ \bottomrule
\end{tabular}
}
\caption{Sample examples from our constructed \pretraindata. The \textcolor{red}{seed words} used to crawl sentences are highlighted in \textcolor{red}{red}. The morality of each sentence is determined by the morality of the corresponding \textcolor{red}{seed word}. For each {\ul moral mention} in text, it is tagged with a special symbol pair, {\ul \textless{}Morality\textgreater{}{}} and {\ul \textless{}/Morality\textgreater{}{}}, and its embodied morality is visually represented using $_{\text{subscript}}$. These morality-bearing example sentences are employed to train \model during \textit{moral word knowledge pre-training stage}.}
\label{tbl:word_example}
\end{table*}

\begin{table*}[t]
\centering
\resizebox{1.0\linewidth}{!}{%
\begin{tabular}{p{0.5\linewidth}llp{0.5\linewidth}p{0.15\linewidth}} \toprule
Scenario                                                                                 & Label                & Scenario Bank            & Label set       & \# of examples                                                                                            \\ \midrule
it is ok to take another person's account and use it as your own.                        & morally disagree     & \texttt{Delphi} (moral agreement) & \{morally agree, morally disagree\}   & $200,000$                                                                      \\
enjoying your life with your family                                                      & morally good         & \texttt{Delphi} (moral judgement) & \{morally good, morally wrong, amoral\}     & $400,000$                                                                 \\
I am working at the local fire station as a fireman. So I should light a lot of matches. & morally unreasonable & \texttt{ETHICS} (deontology)               & \{morally reasonable, morally unreasonable\}  & $18,164$                                                                \\
I usually exercise with my trainer, but stopped because She had a death in her family    & morally reasonable   & \texttt{ETHICS} (justice)                  & \{morally reasonable, morally unreasonable\}  & $21,791$                                                               \\
Wasting your money on something you don't like                                           & morally wrong        & \texttt{Social chem} (judgement)  & \{morally good, morally wrong, amoral\}   & $122,906$                                                                  \\
stay in communication with friends                                                       & loyalty-betrayal     & \texttt{Social chem} (foundation) & \{care-harm, loyalty-betrayal, authority-subversion, fairness-cheating, sanctity-degradation, amoral\}   & $122,906$   \\
faking your relationships                                                                & cheating             & \texttt{Social chem} (morality)   & \{care, harm, loyalty, betrayal, authority, subversion, fairness, cheating, sanctity, degradation, amoral\} & $122,906$ \\ \bottomrule
\end{tabular}
}
\caption{Sample (scenario, label) pairs from our curated suite of Moral Scenario Banks. The seven Moral Scenario Banks are derived from \texttt{Delphi} \citep{jiang2021can}, \texttt{ETHICS} \citep{hendrycks2020aligning} and \texttt{Social Chem} \citep{forbes-etal-2020-social}. Each row represents one scenario bank where the source is listed in \textit{scenario bank} column. \textit{Label set} column shows the full set of plausible labels for each scenario bank. These (Scenario, Label) pairs are employed to train \model during \textit{moral scenario knowledge pre-training stage}. Note, we do not use all data points from \texttt{Delphi} for the sake of training efficiency, but downsample them to the numbers indicated in the the last column \textit{\# of examples}.}
\label{tbl:scenario_example}
\end{table*}

\begin{table*}[t]
\centering
\resizebox{0.75\linewidth}{!}{%
\begin{tabular}{lrrrrrr}
\toprule
                   & \multicolumn{3}{c}{\data}                                                                        & \multicolumn{3}{c}{eMFD Corpus (\citeauthor{hopp2021extended})}                                                \\
\cmidrule(lr){2-4}
\cmidrule(lr){5-7}
                   & \multicolumn{1}{l}{Train}     & \multicolumn{1}{l}{Dev}        & \multicolumn{1}{l}{Test}       & \multicolumn{1}{l}{Train} & \multicolumn{1}{l}{Dev} & \multicolumn{1}{l}{Test} \\ \midrule
\# of stories      & 112                           & 16                             & 30                             & \multicolumn{1}{c}{-}     & \multicolumn{1}{c}{-}    & \multicolumn{1}{c}{-}     \\
\# of articles     & 336                           & 48                             & 90                             & 261                       & 54                      & 96                       \\
\# of sentences    & 9,568                         & 1,256                          & 2,605                          & 10,331                    & 2,042                   & 3,454                    \\
\# of moral events & 4,124                         & 494                            & 876                            & 10,694                    & 1,839                   & 4,513                    \\
\# of moralities & 4,948                         & 606                            & 1,047                          & 11,814                    & 1,958                   & 5,562                    \\
Time range         & \multicolumn{1}{l}{2012-2021} & \multicolumn{1}{l}{01-06/2022} & \multicolumn{1}{l}{07-12/2022} & \multicolumn{1}{c}{-}     & \multicolumn{1}{c}{-}    & \multicolumn{1}{c}{-}   \\ \bottomrule
\end{tabular}
}
\caption{Splits and statistics of \data and eMFD corpus \citep{hopp2021extended}. It is worth noting that a moral event might embody more than one morality.}
\label{tbl:split_stats}
\end{table*}

\begin{table*}[t]
\centering
\resizebox{0.9\linewidth}{!}{%
\begin{tabular}{lrrrrrrrr}
\toprule
                    & \multicolumn{4}{c}{Trigger Detection}                            & \multicolumn{4}{c}{Morality Prediction}                          \\ 
\cmidrule(lr){2-5}
\cmidrule(lr){6-9}
\multicolumn{1}{c}{}                                           & Actual & Intentional & Speculative & \multicolumn{1}{l}{Overall} & Actual & Intentional & Speculative & \multicolumn{1}{l}{Overall} \\ \midrule
Flan-T5 (large)                                                & 47.9          & 44.3          & \textbf{43.1} & 46.2                        & 53.2          & 51.8          & 55.6          & 53.8                        \\
\model                                                           & 50.2          & \textbf{44.4} & 42.1          & 46.9                        & \textbf{56.8} & \textbf{52.4} & \textbf{60.4} & \textbf{57.7}               \\
\;\; w/o moral label association & \textbf{50.3} & 43.9          & \textbf{43.1} & \textbf{47.2}               & 56.6          & \textbf{52.4} & 60            & 57.4  \\ \bottomrule                     
\end{tabular}
}
\caption{F1 results of select models by event status for trigger detection (task B) and morality identification (part of Task C). The performances of the dual-knowledge augmented \model and its variant are based on the aggregated results of \model trained on the three moral scenario banks reported in \Cref{tbl:result_task_BC}. Here, we have observed interesting, substantive findings: 1) \textit{Actual} events are always easier to detect than \textit{non-actual} ones (including both \textit{intentional} and \textit{speculative} events); and 2) In terms of morality prediction, it’s generally easier to predict the morality for \textit{speculative} events. This is attributed to the fact that speculative events are usually presented in the form of ``entity A speculating entity B doing something good/bad to entity C'', which tends to have a higher usage of explicit moral languages.}
\label{tab:event_status}
\end{table*}

\section{Moral Event Schema}
\label{event_schema}
We define a new structured schema for a \textbf{moral event} which represents a moral action.
A moral event encompasses moral agents, moral patients, a morality-bearing event span and event trigger, embodied morality, and event status.
A moral action is performed or enabled by \textbf{moral agents} and affects \textbf{moral patients}.
Moral agents and patients usually possess moral agency, the capability of doing things right or wrong \citep{Gray2009-GRAMTD}, and a moral event may have multiple moral agents and patients.
The \textbf{moral event span} is a contiguous sequence of words in the text that concisely depicts the event/action and carries stand-alone meaning.
This span embodies one or more \textbf{moralities}
in MFT: a moral evaluation will arise when the patient is harmed or helped by the action enabled by the agent \citep{moral_patient, Gray2009-GRAMTD, hopp2021extended}.
Note that moral patients may be \textit{implicit}: they do not have to be mentioned in the \textit{target sentence}.
In line with ACE 2005 \citep{ace2005} and the LDC annotation guideline\footnote{\url{www.ldc.upenn.edu/sites/www.ldc.upenn.edu/files/english-events-guidelines-v5.4.3.pdf}}, the moral event also includes an \textbf{event trigger} that can best characterize the moral action.

To assist the investigation into the linguistic phenomenon of moral events, an event also has an \textbf{event status} which describes the factuality of an event, i.e., whether an event is \textit{actual} or \textit{non-actual} \citep{Saur2009FactBankAC, lee-etal-2015-event}.
We further divide \textit{non-actual} into \textit{intentional} and \textit{speculative} events, where \textit{intentional} describes an event that is being planned or intended to happen, while \textit{speculative} represents an event that may happen, usually speculated by someone who is not an event participant \citep{DemnerFushman2008ThemesIB, Kolhatkar2019TheSO, app12105209}.

\section{\data Annotation Guideline}
\label{annotation_guideline}

\paragraph{Annotation Goal:} Jointly annotate entities (with agency property) and events (with a moral basis).

\paragraph{Entities.}
Entities are the participants in events. They will usually possess moral agency, i.e., the capability of doing things right or wrong \citep{Gray2009-GRAMTD}. There will usually be two entities for every event: the \textbf{agent} is the doer or enabler of the event, and the \textbf{patient} is the one affected by the event.

\textbf{Entity Types:} An entity will often be a Person, Organization, Nation, or something that is backed by entities that have agency.

\textbf{Agency:} Entities usually possess moral agency regardless of whether they are the agent or patient. Sometimes, an entity itself might not have agency but is backed by some other entities that have agency. For example, ``hurting the Constitution'' essentially means ``hurting the people''. The Constitution itself has no agency, but the people behind the Constitution have agency, so we annotate either ``Constitution'' or ``People'' as the moral patient.

\textbf{Canonical Names} are uniquely identified strings in a knowledge base such as Wikipedia. Entities should be annotated with their canonical names, if possible. An entity's canonical name might not be the first occurrence of that name in the article. For consistency, please use the same canonical name throughout the entire article. For example, mentions of ``President Trump'' or ``Trump'' should be annotated as ``Donald Trump''.

\paragraph{Moral Events.}
Moral events have a basis in moral foundations and possess moral evaluations that arise when the patient has agency and can be harmed or helped by an action/event \citep{moral_patient, Gray2009-GRAMTD}. The annotated event must be a concise span that exactly appears in the text, and it should carry stand-alone meaning.

\textbf{Event Entities:}
Agent \& Patient: For each moral event, there must be at least one enabler (agent) as well as at least one affected (patient). If the agent and patient are not apparent in the text, please infer them to make sure both agent and patient are present. For example, in the following sentence ``That briefing averted congressional criticism, even though the administration formally missed a deadline to implement sanctions targeting Russian defense and intelligence industries'',  we can tell that there is a moral event ``missed a deadline'' (which embodies a morality of betrayal), and the associated agent is ``Trump Administration''. However, the patient is not explicitly stated, but we can infer ``Congress'' as a patient since missing the deadline would impede Congress from implementing sanctions or taking further actions.

\textbf{Moral Foundations:} Follow MFT \citep{Graham2009LiberalsAC}, MFD 2.0 \citep{frimer2019moral} and supplementary materials of eMFD \citep{hopp2021extended} to annotate the moral foundation(s) embodied in each moral event. Note, a moral event can embody more than one morality.
\begin{itemize}
\item[1.] \textbf{Ten moralities:} There are five moral foundations, each with a positive and negative polarity: Care, Harm, Fairness, Cheating, Loyalty, Betrayal, Authority, Subversion, Sanctity, and Degradation.  

\item[2.] \textbf{Author’s Point of View:} During the annotation process, \textit{annotate from the author’s perspective rather than the audience's}. In other words, consider what the author is trying to say or imply by writing these words. You may also consider why the author included this event, and what kind of morality is embodied through the inclusion of this event.

\item[3.] \textbf{Morality Toward the Patient:} The annotated morality should reflect the perception of the patients, towards whom an agent performs a moral action.
\end{itemize}

\noindent Additionally, we also annotate the following \textbf{Event Status} to reveal the linguistic construct of a moral event.

\textbf{Event Status:} An event has one of three statuses:
\begin{itemize}
\item Actual: An event that is happening or has happened.
\item Intentional: An event that is being planned or intended to happen in the future. Usually, it is the moral agent’s subjective intention of the event.
\item Speculative: An event that may happen, usually speculated by someone who is not a participant in the event (e.g. the speaker of a quote, or the author of the article). This can be used to mark an unsubstantiated guess of a past/current/future event.
\end{itemize}

\section{\data Annotation Quality Control}
\label{data_quality_control}

We ensured the quality of the annotations at multiple steps in the collection process.
All annotators participated in a training phase before beginning the annotations.
In addition, the annotators participated in a weekly review session with the authors who would answer questions and provide guidance for annotators to revise their annotations.

We also found high inter-annotator agreement. {This paragraph is based on comparing article annotations before and after the \textbf{revision}, i.e., approach (a) as described in~\Cref{annotation_process}.}
To compute agreement, we first identify overlapping moral event text spans where half of the words are identical, and then obtain Krippendorff's alpha's on the annotated properties (e.g., Agent, Patient) of the events.  Agreement levels are included in~\cref{tab:agreement}.
The revised articles have on average 5.7\% more annotations than the first-pass articles.
In terms of the nature of disagreements, some disagreements were on whether an event was negated.
For example, a sentence like ``the president did not sign the bill'' contains a clearly negated event, due to the presence of the word ``not.''
However, in the sentence ``the president hesitated to sign the bill'', one annotator could have annotated the event ``hesitated'', while another could annotate the negated event ``sign''.
In addition, annotators sometimes disagreed on the morality of an event.
For example, ``the Supreme Court overrule the case'' could be marked as Harm towards one patient, or Care towards a different patient.
Many of these such annotations are subjective, though overall we find that these disagreements do not substantially lower the quality of our dataset.
For this project, we use the revised annotations as training and testing data for our models.

{Likewise, a similar quality control study is conducted on annotated articles undergoing \textbf{merging}, approach (b) as described in ~\cref{annotation_process}. Agreement levels are included in~\cref{tab:agreement}. For this portion of data, we use the merged annotations as training and testing data for our models.
Agreement on the article's ideological leaning is 0.7577. }

{Furthermore, upon comparing all annotated articles, our annotated \textbf{article leanings} match AllSides' media-level labels for $70.9\%$ and $76.4\%$  of the time before and after the second-pass adjudication, respectively. We follow \citet{seesaw} and consider the difference between our annotated article leaning and AllSides label within one level as a match, e.g., Left (0) and Lean Left (1) are matched.
This further illuminates the high quality of \data and the effectiveness of our designed two-pass annotation process.}

\begin{table}[t]
\centering
\resizebox{0.8\linewidth}{!}{%
\begin{tabular}{lrrr}
\toprule
                     & Virtue & Vice  & Proportion \\
\midrule
Care/Harm            & 1,348  & 2,060 & 51.6\%      \\
Fairness/Cheating    & 531    & 453   & 14.9\%      \\
Loyalty/Betrayal     & 329    & 257   & 8.9\%       \\
Authority/Subversion & 1,140  & 418   & 23.6\%      \\
Sanctity/Degradation & 19     & 46    & 1.0\%       \\ \midrule
Total                     & 3,367  & 3,234 & 100.0\%    \\
\bottomrule
\end{tabular}
}
\caption{
Distribution of moralities in moral event annotations in \data. 
Numbers in \textit{Virtue} and \textit{Vice} columns are raw counts of annotated moralities.
}
\label{tbl:distribution}
\end{table}

\begin{table}[t]
\centering
\resizebox{0.95\linewidth}{!}{%
\begin{tabular}{lcccc}
\toprule
\multicolumn{1}{c}{\multirow{2}{*}{Model}}                                                     & \multicolumn{2}{c}{\data}                    & \multicolumn{2}{c}{eMFD Corpus} \\
\cmidrule(lr){2-3} 
\cmidrule(lr){4-5}
\multicolumn{1}{c}{}                                                                           & F1                   & Acc.                 & F1             & Acc.           \\ \midrule
\multicolumn{5}{l}{\textbf{Baselines}}                                                                                                                                         \\
Dictionary-based counting (\citeauthor{brady2017emotion})                                                                        & 45.8                 & 56.8                 & 33.0           & 52.0           \\
RoBERTa-large (large; \citeauthor{Liu2019RoBERTaAR})                                                                                   & 63.6                 & 82.6                 & 28.7           & 69.0           \\
POLITICS  (base; \citeauthor{politics})                                                                                      & 62.7                 & 82.4                 & 29.0           & 68.8           \\
ChatGPT (zero-shot; \citeauthor{Li2023EvaluatingCI})                                                                              & 41.2                 & 69.9                 & 31.9           & 66.9           \\
ChatGPT (few-shot; \citeauthor{Li2023EvaluatingCI})                                                                             & 46.9                 & 75.6                 & 30.5           & 69.1           \\
Flan-T5 (large; \citeauthor{Chung2022ScalingIL})                                                                                 & 62.0                 & 83.6                 & 25.4           & 68.4           \\ \midrule
\multicolumn{5}{l}{\textbf{MOKA with moral word knowledge augmentation only}}                                                                                                  \\
Pretrain on \texttt{Morality Bank} only                                                                            & 63.6                 & \reda{83.9}                 & 27.3           & 69.0           \\
\;\; + moral word linking ($\mathcal{L}_{MWL}$)                                       & \reda{63.9}                 & \reda{83.9}                 & 27.8           & 69.0           \\
\;\;\;\; + moral label association ($\mathcal{L}_{MLA}$) & \redb{64.0}                 & \reda{83.9}                 & 28.5           & 69.1           \\ \midrule
\multicolumn{5}{l}{\textbf{MOKA with moral scenario knowledge augmentation only}}                                                                                              \\
Delphi (moral agreement; \citeauthor{jiang2021can})                  & 62.5 & \reda{84.0} & 30.0 & \redb{70.2} \\
\;\; + RLM & 63.2 & \redb{84.2} & 30.3 & \redb{70.3} \\
Delphi (moral judgement; \citeauthor{jiang2021can})                                                                       & \reda{63.7}                 & \redb{84.1}                 & 30.4           & \redb{70.4}           \\
\;\; + RLM                                                      & 62.3                 & \reda{83.8}                 & 30.1           & \redb{70.3}           \\
Deontology  (\citeauthor{hendrycks2020aligning})                                                                                   & 62.5                 & 83.6                 & 30.5           & \redc{70.5}           \\
\;\; + RLM                                                      & 62.2                 & 83.5                 & 30.4           & \redb{70.4}           \\
Justice  (\citeauthor{hendrycks2020aligning})                                    & 62.5    & \reda{83.7}   & 30.4   & \redb{70.3}   \\
\;\; + RLM    & 62.4    & 83.6   & 31.8   & \redc{70.6}   \\
Social chem (judgement; \citeauthor{forbes-etal-2020-social})                      & 63.6    & \redb{84.2}   & 30.0   & \redb{70.2}   \\
\;\; + RLM    & 62.9    & 83.6   & 30.7   & \redb{70.4}   \\
Social chem (foundation; \citeauthor{forbes-etal-2020-social})                                                                        & 62.2                 & \reda{83.7}                 & 32.4           & \redc{70.6}           \\
\;\; + RLM                                                      & \redb{64.1}                 & \reda{84.0}                 & 32.5           & \redc{70.7}           \\ 
Social chem (morality; \citeauthor{forbes-etal-2020-social})                       & 62.7    & \reda{83.8}   & 30.0   & \redb{70.3}   \\
\;\; + RLM    & 63.3    & \redb{84.1}   & 32.5   & \redc{70.6}  \\ \midrule
\multicolumn{5}{l}{\textbf{MOKA with dual moral knowledge augmentation}}                                                                                                  \\
Delphi (moral agreement; \citeauthor{jiang2021can})                                       & \redb{64.4}          & \redb{84.3}          & \redc{34.0}          & \redc{71.0}          \\
\;\; - $\mathcal{L}_{MLA}$ & 63.3          & \reda{84.0}          & \redb{33.2}          & \redc{70.8}          \\
Delphi (moral judgement; \citeauthor{jiang2021can})                                                                       & 63.3                 & 83.6                 & 32.9           & \redc{70.7}           \\
\;\; - $\mathcal{L}_{MLA}$                                 & \reda{63.9}                 & \redb{84.1}                 & 32.1           & \redc{70.6}           \\
Deontology (\citeauthor{hendrycks2020aligning})                                                                                     & \redb{64.0}                 & \reda{84.0}                 & 32.9           & \redc{70.8}           \\
\;\; - $\mathcal{L}_{MLA}$                                 & \redb{64.2}                 & \reda{84.0}                 & \redc{34.3}  & \redc{71.1}  \\
Justice  (\citeauthor{hendrycks2020aligning})                                                       & \redb{64.0}          & \reda{84.0}          & 32.9          & \redc{71.0}          \\
\;\; - $\mathcal{L}_{MLA}$ & \reda{63.7}          & \redb{84.1}          & \redb{33.3}          & \redc{71.0}          \\
Social chem (judgement; \citeauthor{forbes-etal-2020-social})                                        & \redb{64.3}          & \redb{84.2}          & 32.7          & \redc{70.9}          \\
\;\; - $\mathcal{L}_{MLA}$ & \redb{64.2}          & \redb{84.3}          & \redb{33.4}          & \redc{71.1}          \\
Social chem (foundation; \citeauthor{forbes-etal-2020-social})                                                                        & \redc{\textbf{65.3}}        & \redb{\textbf{84.3}}        & \redc{33.7}           & \redc{71.0}           \\
\;\; - $\mathcal{L}_{MLA}$                                 & \redb{64.1} & \reda{84.0} & \redb{33.4}           & \redc{71.0}           \\ 
Social chem (morality; \citeauthor{forbes-etal-2020-social})                                         & \redb{64.5}          & \reda{83.9}          & \redd{\textbf{34.6}} & \redc{\textbf{71.3}} \\
\;\; - $\mathcal{L}_{MLA}$ & \reda{63.8}          & \reda{84.0}          & \redb{33.3}          & \redc{70.9}  \\ \midrule
Improvements over best baseline                                                                & 2.7\%                & 0.8\%                & 4.8\%          & 3.2\%   \\
\bottomrule
\end{tabular}
}
\caption{
Full weighted F1 and accuracy results on \data and eMFD Corpus \citep{hopp2021extended} for task A (average of 5 runs). 
Best results are in \textbf{bold}. 
``+'' and ``-'' indicate the inclusion or exclusion of a particular training objective.
\textit{Color scheme}: \model and its single-knowledge-augmentation variants are highlighted on a scale of 5 red shades based on the relative improvements over the strongest baseline. They are highlighted in \reda{pale pink},  \redb{pink}, \redc{rose-pink}, \redd{rose-red} and \rede{dark red}, if the relative gains are in the range of $(0.0\%-0.5\%]$, $(0.5\%-2.0\%]$, $(2.0\%-4.0\%]$, $(4.0\%-7.0\%]$ and $(7.0\%-\infty\%)$, respectively.
}
\vspace{-2mm}
\label{tbl:full_result_task_A}
\end{table}

\begin{table*}[t]
\centering
\resizebox{0.9\linewidth}{!}{%
\begin{tabular}{lcccccc}
\toprule
\multicolumn{1}{c}{\multirow{2}{*}{Model}}                                                     & \multicolumn{1}{c}{Task B} & \multicolumn{5}{c}{Task C}                                                    \\ 
\cmidrule(lr){2-2} 
\cmidrule(lr){3-7}
\multicolumn{1}{c}{}                                                                           & Trigger EM                 & Morality F1      & Agent EM      & Agent F1      & Patient EM    & Patient F1    \\ \midrule
\multicolumn{7}{l}{\textbf{Baselines}}                                                                                                                                                                               \\
DEGREE (base; \citeauthor{hsu-etal-2022-degree})                                                                                  & 45.5                       & 53.0          & 47.3          & 58.6          & 30.1          & 39.2          \\
DEGREE (large; \citeauthor{hsu-etal-2022-degree})                                                                                 & 46.2                       & 54.2          & 49.2          & 60.3          & 30.5          & 40.3          \\
ChatGPT (zero-shot; \citeauthor{Li2023EvaluatingCI})                                                                            & 19.5                       & 39.5          & 30.3          & 49.8          & 12.3          & 23.2          \\
ChatGPT (few-shot; \citeauthor{Li2023EvaluatingCI})                                                                             & 32.1                       & 38.1          & 34.2          & 51.4          & 20.1          & 30.6          \\
Flan-T5 (large; \citeauthor{Chung2022ScalingIL})                                                                                & 46.2                       & 53.8          & 47.5          & 59.4          & 30.8          & 41.2          \\ \midrule
\multicolumn{7}{l}{\textbf{MOKA with moral word knowledge augmentation only}}                                                                                                                               \\
Pretrain on \texttt{Morality Bank} only                                                                            & 45.3                       & \redb{54.6}          & 47.5          & 59.9          & \redb{31.2}          & \redb{41.7}          \\
\;\; + moral word linking ($\mathcal{L}_{MWL}$)                                       & 45.6                       & \redc{55.9}          & 47.6          & 59.8          & \redc{\textbf{31.5}} & \redb{41.7}          \\
\;\;\;\; + moral label association ($\mathcal{L}_{MLA}$) & 46.2                       & \redd{57.0}          & 48.3          & 60.2          & \redb{31.3}          & \redb{41.9}          \\ \midrule
\multicolumn{7}{l}{\textbf{MOKA with moral scenario knowledge augmentation only}}                                                                                                                           \\
Delphi (moral agreement; \citeauthor{jiang2021can})                  & \redb{46.6} & \redc{55.9} & 48.9          & \redb{60.9}          & 30.8          & \redb{41.5}          \\
\;\; + RLM & \redc{47.6} & \redc{56.3} & 48.6          & \reda{60.5}          & \redc{\textbf{31.6}} & \redb{41.8}          \\
Delphi (moral judgement; \citeauthor{jiang2021can})                                                                       & \redb{47.0}                       & \redd{57.5}          & 48.5          & \reda{60.4}          & \reda{30.9}          & \reda{41.4}          \\
\;\; + RLM                                                                                          & \redc{47.4}                       & \redc{55.6}          & 48.5          & 60.3          & \redb{31.2}          & \redb{41.5}          \\
Deontology (\citeauthor{hendrycks2020aligning})                                                                                     & 46.1                       & \redb{54.8}          & 49.0          & \redb{60.9}          & \reda{30.9}          & \redb{41.6}          \\
\;\; + RLM                                                                                          & \redc{47.2}                       & \redc{56.0}          & \redb{\textbf{49.5}} & \redb{61.2}          & \redb{31.3}          & \redc{\textbf{42.1}} \\
Justice (\citeauthor{hendrycks2020aligning})                                   & \redb{46.6} & \redb{54.7} & 48.7          & \redb{60.7}          & \redb{31.0}          & \redb{41.5}          \\
\;\; + RLM & \redb{46.9} & \redb{55.2} & 48.6          & \redb{60.8}          & \redb{31.4}          & \redb{41.6}          \\
Social chem (judgement; \citeauthor{forbes-etal-2020-social})                   & \redb{47.1} & \redc{55.4} & 48.6          & \redb{60.9}          & \redb{31.2}          & 41.2          \\
\;\; + RLM & \redc{47.2} & \redb{54.9} & 48.5          & 60.1          & \redb{31.3}          & \redb{41.6}          \\
Social chem (foundation; \citeauthor{forbes-etal-2020-social})                                                                       & \redb{46.7}                       & \redd{56.5}         & 48.9          & \redb{\textbf{61.4}} & \redb{31.0}          & \reda{41.4}          \\
\;\; + RLM                                                                                          & \redc{47.5}                       & \redc{56.0}          & 48.8          & \reda{60.5}          & \redb{31.0}          & \redb{41.7}          \\ 
Social chem (morality; \citeauthor{forbes-etal-2020-social})                    & \redb{46.8} & \redc{56.3} & 48.6          & \reda{60.6}          & \redb{31.2}          & 40.7          \\
\;\; + RLM & \redc{47.2} & \redc{55.5} & 48.7          & \redb{60.7}          & \redb{31.0}          & \redb{41.5}         \\ \midrule
\multicolumn{7}{l}{\textbf{MOKA with dual moral knowledge augmentation}}                                                                                                                               \\
Delphi (moral agreement; \citeauthor{jiang2021can})       &  \redb{46.5} & \redd{56.9} & 48.4 & \reda{60.5} & 30.5 & 41.0 \\                               
\;\; - $\mathcal{L}_{MLA}$ &  \redc{47.3} & \redd{57.2} & 47.5 & \reda{60.6} & 30.7 & 40.9 \\
Delphi (moral judgement; \citeauthor{jiang2021can})                                                                       & \redc{47.4}                       & \redd{56.8}          & 48.1          & 60.3          & 30.2          & 40.5          \\
\;\; - $\mathcal{L}_{MLA}$                                                                      & \redb{46.7}                       & \redd{57.2}          & 47.6          & 60.0          & 30.2          & 40.5          \\
Deontology (\citeauthor{hendrycks2020aligning})                                                                                       & \redb{46.8}                       & \rede{58.2}         & 47.9          & 60.3          & \reda{30.9}          & 41.1          \\
\;\; - $\mathcal{L}_{MLA}$                                                                      & \redd{\textbf{48.1}}              & \redd{57.3}          & 48.2          & \redb{61.0}          & 30.7          & 41.1          \\
Justice  (\citeauthor{hendrycks2020aligning})                                                         & \redb{46.9} & \redd{57.4} & 48.6 & \redb{61.1} & \redb{31.0} & 41.2 \\
\;\; - $\mathcal{L}_{MLA}$ & \redc{47.4} & \redd{56.9} & 48.0 & \redb{60.9} & \redb{31.1} & 41.1  \\
Social chem (judgement; \citeauthor{forbes-etal-2020-social})                                        &  \redb{46.5} & \redd{57.7} & 47.9 & \reda{60.6} & 29.7 & 40.6 \\
\;\; - $\mathcal{L}_{MLA}$ &  \redb{46.8} & \redd{57.7} & 47.5 & \redb{60.9} & 30.1 & 40.3 \\
Social chem (foundation; \citeauthor{forbes-etal-2020-social})                                                                        & \redb{46.5}                       & \rede{58.1} & 48.4          & \redb{61.0}         & 30.5          & 40.8          \\
\;\; - $\mathcal{L}_{MLA}$                                                                      & \redb{46.7}                       & \redd{57.7}          & 48.2          & \reda{60.5}          & 30.0          & 40.1          \\ 
Social chem (morality; \citeauthor{forbes-etal-2020-social})                                         &  \redb{47.0} & \rede{58.2} & 48.0 & \reda{60.6} & 30.5 & 40.5 \\
\;\; - $\mathcal{L}_{MLA}$ & \redb{46.7} & \rede{\textbf{58.5}}  & 47.9 & \redb{60.8} & 30.2 & 40.9 \\ \midrule
Improvements over best baseline                                                                                  & 4.1\%                      & 7.9\%         & 0.6\%         & 1.8\%         & 2.6\%         & 2.2\%             \\
\bottomrule
\end{tabular}
}
\caption{
Full results on \data for tasks B and C, and breakdown of performances by event attributes (average of 5 runs). 
Best results are in \textbf{bold}.
``+'' and ``-'' indicate the inclusion or exclusion of a particular training objective.
\textit{Color scheme}: \model and its single-knowledge-augmentation variants are highlighted on a scale of 5 red shades based on the relative improvements over the strongest baseline. They are highlighted in \reda{pale pink},  \redb{pink}, \redc{rose-pink}, \redd{rose-red} and \rede{dark red}, if the relative gains are in the range of $(0.0\%-0.5\%]$, $(0.5\%-2.0\%]$, $(2.0\%-4.0\%]$, $(4.0\%-7.0\%]$ and $(7.0\%-\infty\%)$, respectively.
}
\label{tbl:full_result_task_BC}
\end{table*}

\begin{table*}[t]
\centering
\resizebox{0.88\linewidth}{!}{%
\begin{tabular}{lll}
\toprule
Task   & Text                                                                                                          &  \\ \midrule
Task A & \begin{tabular}[c]{@{}l@{}}Moral Event Detection task definition:\textbackslash{}n\textbackslash\\ Given an input list of words from a news article, identify the moral event trigger in the input list. An event \textbackslash\\ is something that happens, a specific occurrence involving participants, and can frequently be described as a change of state. \textbackslash\\ A moral event has a basis in moral foundations, and possesses moral evaluations which arise when the patient has agency \\ and can be harmed or helped by an action undertaken by an agent. \textbackslash\\ A moral event trigger is the main word or phrase that most explicitly \textbackslash\\ expresses the occurrence of a moral event.\textbackslash{}n\textbackslash{}n\textbackslash\\ In the input list, special tokens are defined as follows. \textbackslash\\ \textless{}Title\textgreater and \textless{}/Title\textgreater enclose the title of the news article; \\ \textless{}News\textgreater and \textless{}/News\textgreater enclose the truncated content of the news article; \textless{}Target\textgreater and \textless{}/Target\textgreater \\ enclose the target sentence from which the event trigger should be extracted. \textbackslash{}n\textbackslash\\ The output of the Moral Event Detection task should be a dictionary in the json format. Each \textbackslash\\ dictionary corresponds to a trigger and should consist of \textbackslash{}"trigger\textbackslash{}", \textbackslash{}"start\_word\_index\textbackslash{}", \textbackslash\\ \textbackslash{}"end\_word\_index\textbackslash{}", \textbackslash{}"confidence\textbackslash{}" four keys. The value of \textbackslash{}"start\_word\_index\textbackslash{}" key and \textbackslash{}"end\_word\_index\textbackslash{}" key are the \textbackslash\\ index (zero-indexed) of the start and end word of \textbackslash{}"trigger\textbackslash{}" in the input list, respectively. The \textbackslash\\ value of \textbackslash{}"confidence\textbackslash{}" key is an integer ranging from 0 to 100, indicating how confident you are that \textbackslash\\ the \textbackslash{}"trigger\textbackslash{}" expresses a moral event. \textbackslash\\ Note that your answer should only contain the json string and nothing else.\textbackslash{}n\textbackslash{}n\textbackslash\\ You will first see 5 demonstrations of the task, and then you will be asked to perform the task for a given input list.\textbackslash{}n\textbackslash{}n\\ \\ Demonstration i: \textless{}Demostration i\textgreater\\ \\ \textbackslash{}nPerform Moral Event Detection task for the following input list, and print the output:\textbackslash{}n\\ \\ {[}“This”, “is”, “a”, “sample”, “input”{]}\end{tabular}                                                                                                                                          &  \\ \midrule
Task B & \begin{tabular}[c]{@{}l@{}}Moral Dimension Prediction definition:\textbackslash{}n\textbackslash\\ Given a moral event span and an input list of words from a news article, make a 5-way judgment on the moral dimension for the given moral event. \textbackslash\\ A more event span might embody more than one moral dimension. An event \textbackslash\\ is something that happens, a specific occurrence involving participants, and can frequently be described as a change of state. \textbackslash\\ A moral event has a basis in moral foundations, and possesses moral evaluations which arise when the patient has agency \\ and can be harmed or helped by an action undertaken by an agent. \textbackslash\\ The five moral dimensions are 'Care/Harm', 'Fairness/Cheating', 'Loyalty/Betrayal', 'Authority/Subversion', and 'Sanctity/Degradation'\textbackslash{}n\textbackslash{}n\textbackslash\\ In the input list, special tokens are defined as follows: \textbackslash\\ \textless{}Title\textgreater and \textless{}/Title\textgreater enclose the title of the news article; \textless{}News\textgreater and \textless{}/News\textgreater enclose the truncated content of the news article; \textbackslash\\ \textless{}Target\textgreater and \textless{}/Target\textgreater enclose the target sentence where the target moral event span stands; \textless{}Event\textgreater and \textless{}/Event\textgreater enclose the target moral event span.\textbackslash{}n\textbackslash\\ The output of the Moral Event Detection task should be a dictionary in the json format. Each \textbackslash\\ dictionary corresponds to a moral event and should consist of \textbackslash{}"moral dimensions\textbackslash{}" and \textbackslash{}"confidence\textbackslash{}" two keys. \\ The value of \textbackslash{}"moral dimensions\textbackslash{}" should be a list of predicted moral dimensions that are embodied in the target moral event span. \textbackslash\\ The value of \textbackslash{}"confidence\textbackslash{}" key is an integer ranging from 0 to 100, indicating how confident you are that \textbackslash\\ the moral event span embodies predicted \textbackslash{}"moral dimensions\textbackslash{}". \textbackslash\\ Note that your answer should only contain the json string and nothing else.\textbackslash{}n\textbackslash{}n\textbackslash\\ You will first see 5 demonstrations of the task, and then you will be asked to perform the task for a given input list. \textbackslash{}n\textbackslash{}n\\ \\ Demonstration i: \textless{}Demostration i\textgreater\\ \\ \textbackslash{}nPerform Moral Dimension Prediction task for the following input list, and print the output:\textbackslash{}n\\ \\ {[}“This”, “is”, “a”, “sample”, “input”{]}\end{tabular}                                                                                                                                                   &  \\ \midrule
Task C & \begin{tabular}[c]{@{}l@{}}Moral Event Argument Extraction task definition:\textbackslash{}n\textbackslash\\ Given an input list of words from a news article and a moral event span, identify moral event arguments for the given moral event span. \textbackslash\\ Specifically, moral event arguments consists of three attributes: moral agent, moral patient and 10-way morality prediction. \textbackslash\\ An event is something that happens, a specific occurrence involving participants, and can frequently be described as a change of state. \textbackslash\\ A moral event has a basis in moral foundations, and possesses moral evaluations which arise when the patient has agency \\ and can be harmed or helped by an action undertaken by an agent. \textbackslash\\ A moral event span is a main word or phrase that most explicitly \textbackslash\\ expresses the occurrence of a moral event. A moral agent is the doer or enabler of a moral event, \\ and the moral patient is the one affected by the moral event. \textbackslash\\ The ten moralities are 'Care', 'Harm', 'Fairness', 'Cheating', 'Loyalty', 'Betrayal', 'Authority', 'Subversion', 'Sanctity', and 'Degradation'\textbackslash{}n\textbackslash{}n\textbackslash\\ In the input list, special tokens are defined as follows. \textbackslash\\ \textless{}Title\textgreater and \textless{}/Title\textgreater enclose the title of the news article; \textless{}News\textgreater and \textless{}/News\textgreater enclose the truncated content of the news article; \textbackslash\\ \textless{}Target\textgreater and \textless{}/Target\textgreater enclose the target sentence where the target moral event span stands; \textless{}Event\textgreater and \textless{}/Event\textgreater enclose the target moral event span.\textbackslash{}n\textbackslash\\ The output of the Moral Event Argument Extraction task should be a dictionary in the json format. Each \textbackslash\\ dictionary corresponds to a moral event span and should consist of \textbackslash\\ \textbackslash{}"agent\textbackslash{}", \textbackslash{}"confidence-agent\textbackslash{}", \textbackslash\\ \textbackslash{}"patient\textbackslash{}", \textbackslash{}"confidence-patient\textbackslash{}", \textbackslash\\ \textbackslash{}"morality\textbackslash{}" and \textbackslash{}"confidence-value\textbackslash{}" six keys. \textbackslash{}n\textbackslash\\ The value of \textbackslash{}"agent\textbackslash{}" and \textbackslash{}"patient\textbackslash{}" keys should be a list of moral agents and moral patients in their canonical names, respectively. \textbackslash\\ Note, canonical names are uniquely-identified strings in a knowledge base such as Wikipedia. \\ An entity's canonical name might not be explicitly mentioned in the input list. \textbackslash\\ For example, the canonical names of \textbackslash{}"Trump\textbackslash{}", \textbackslash{}"Republican\textbackslash{}", \textbackslash{}"Democrats\textbackslash{}", \textbackslash{}"Senate\textbackslash{}", and \textbackslash{}"United States Department of State\textbackslash{}" are \\ \textbackslash{}"Donald Trump\textbackslash{}", \textbackslash{}"Republican Party\textbackslash{}", \textbackslash{}"Democratic Party\textbackslash{}", \textbackslash{}"United States Senate\textbackslash{}", and \textbackslash{}"State Department\textbackslash{}", respectively. \textbackslash{}n\textbackslash\\ The value of \textbackslash{}"confidence-agent\textbackslash{}" key is an integer ranging from 0 to 100, indicating how confident you are that \textbackslash\\ the value of \textbackslash{}"agent\textbackslash{}" key plays the agent role in the target moral event. \textbackslash{}n\textbackslash\\ The value of \textbackslash{}"confidence-patient\textbackslash{}" key is an integer ranging from 0 to 100, indicating how confident you are that \textbackslash\\ the value of \textbackslash{}"patient\textbackslash{}" key plays the patient role in the target moral event. \textbackslash{}n\textbackslash\\ The value of \textbackslash{}morality\textbackslash{}" should be a list of predicted moralities that are embodied in the target moral event span. \textbackslash\\ The value of \textbackslash{}"confidence-value\textbackslash{}" key is an integer ranging from 0 to 100, indicating how confident you are that \textbackslash\\ the moral event span embodies predicted \textbackslash{}"moralities\textbackslash{}". \textbackslash{}n\textbackslash\\ Note that your answer should only contain the json string and nothing else.\textbackslash{}n\textbackslash{}n\textbackslash\\ You will first see 5 demonstrations of the task, and then you will be asked to perform the task for a given input list. \textbackslash{}n\textbackslash{}n\\ \\ Demonstration i: \textless{}Demostration i\textgreater\\ \\ \textbackslash{}nPerform Moral Dimension Prediction task for the following input list, and print the output:\textbackslash{}n\\ \\ {[}“This”, “is”, “a”, “sample”, “input”{]}\end{tabular} & \\ \bottomrule
\end{tabular}
}
\caption{Prompts used to test ChatGPT's moral reasoning capability, adapted from \citet{Li2023EvaluatingCI}. For task A, although we prompt ChatGPT to predict \textit{start} and \textit{end} indexes in its structural output, we only use its predicted value of the \textit{trigger} field, due to ChatGPT’s insufficient numerical reasoning capabilities.}
\label{tbl:chatgpt_prompt}
\end{table*}

\section{Implementation Details of Lexicon Memory Access}
\label{appx:lma}
\paragraph{Lexicon Memory Access.} 
Access to the Lexicon Memory is triggered when encountering the morality special tokens as follows.
\model takes as a query vector $\mathbf{h}_q$, the averaged representation of the special token pair \texttt{(<Morality>, </Morality>)} and the moral mention in between. $\mathbf{h}_q$ is then used to retrieve relevant moral knowledge $\mathbf{h}_m$ from the Lexicon Memory via a single-head attention mechanism.

\begin{align}  
    \mathbf{h}_m =& \mathbf{W}_{2}(\Sigma \alpha_i \cdot \mathbf{m}_i) \\
    \alpha_i=& \frac{\exp(\mathbf{m}_i^\top \mathbf{W}_1 \mathbf{h}_q)}{\Sigma_{\mathbf{m}_j \in \mathcal{M}} \exp(\mathbf{m}_j^\top \mathbf{W}_1 \mathbf{h}_q)} \label{appx:eq:alpha}
\end{align}

\noindent where $\mathcal{M}$  denotes the morality lexicon, $\mathbf{m}_i$ is a moral word embedding, and $\mathbf{W}_1$ and $\mathbf{W}_2$ are learnable matrices. Eventually, $\mathbf{h}_m$ is added to $\mathbf{h}_q$, the sum of which is normalized before being fed to the next Transformer layer, which is $9^{\text{th}}$ layer in \model encoder.

\paragraph{Moral Word Knowledge Pre-training.} 
The pretraining objective is a combination of language modeling ($\mathcal{L}_{LM}$), morality prediction ($\mathcal{L}_{MV}$), and moral word linking ($\mathcal{L}_{MWL}$) and moral label association ($\mathcal{L}_{MLA}$). In this part, we provide detailed mathematical formulations for $\mathcal{L}_{MWL}$ and $\mathcal{L}_{MLA}$.

Without loss of generality, the input sentence is defined as $\mathbf{x} = [x_1, x_2,\cdot\cdot\cdot, x_T]$ of length $T$ which contains a set of moral mentions $\{(t_i, m_i, V_i)\}$, where $t_i = x_j$ for some $j \in [1, T]$. Here, $t_i$ is a moral mention in $\mathbf{x}$, $m_i$ is the corresponding moral word, and $V_i = \{v_{i,1}, v_{i,2}, ...\}$ is the set of associated moralities.

$\mathcal{L}_{MWL}$: For each moral mention ($t_i$) in text, the moral word linking objective guides \model to identify the corresponding moral word ($m_i$) by learning to \textbf{maximize} the attention score over the correct entry. That is, \textbf{maximize} for $\mathcal{L}_{MWL} \eqdef \alpha_{m_i}$, where $\alpha_{m_i}$ is computed using \cref{appx:eq:alpha}.

$\mathcal{L}_{MLA}$: For each moral mention ($t_i$) in text, the moral label association objective is, for each morality ($v_{i,k}$) embodied by the mention, maximize the summation of attention scores over all moral words that share the same morality. Here, we denote $M_{v_{i,k}}$ as a set of moral words defined in $\mathcal{M}$ that carry $v_{i,k}$ value, where $M_{v_{i,k}} \subset \mathcal{M}$. We compute the aggregated  attention score ($\mathrm{A}_{i,k}$) for each embodied morality as follows:
\begin{align} 
   \mathrm{A}_{i,k} &= \sum_{m_p\in M_{v_{i,k}}}\alpha_{m_p}
\end{align}

\noindent where $\alpha_{m_p}$ is computed using \cref{appx:eq:alpha}. We then denote $\mathbf{A}_{i}$ as the set of embodied moralities' aggregated attention scores, i.e., $\mathbf{A}_{i} = \{\mathrm{A}_{i,1}, \mathrm{A}_{i,2}, ...\}$ where $|\mathbf{A}_{i}|= |V_i|$. For simplicity, we use $\mathbf{A}_{i}^C$ to represent the complement set, i.e., a set of aggregated attention scores of non-embodied moralities. To support the training of moral words that might be associated with more than one morality, we \textbf{minimize} \textit{multi-label margin loss} as shown in \cref{appx:eq:lmla}:
\begin{align} 
   \mathcal{L}_{MLA} &\eqdef \frac{\sum_{y\in \mathbf{A}_{i}}  \sum_{z\in \mathbf{A}_{i}^C} \max(0, 1-(y-z)) }{|\mathbf{A}_{i}| + |\mathbf{A}_{i}^C|} \label{appx:eq:lmla} \\
   &\eqdef \frac{\sum_{y\in \mathbf{A}_{i}}  \sum_{z\in \mathbf{A}_{i}^C}  (1 + z - y) } {10} \label{appx:eq:lmla_simplified}
\end{align}

We derive \cref{appx:eq:lmla_simplified} from \cref{appx:eq:lmla}, since we notice that aggregated attention scores are always bound by $[0,1]$, and there is a fixed number of plausible moralities, which is 10.

\end{document}